\documentclass[12pt]{article}
\usepackage{xr}
\usepackage[margin=1.0in]{geometry}
\usepackage{caption}

\usepackage[T1]{fontenc}
\usepackage[utf8]{inputenc}
\usepackage{authblk}
\usepackage[numbers]{natbib}

\title{Stick-Breaking Policy Learning in Dec-POMDPs}

\author[1]{Miao Liu}
\author[2]{Christopher Amato}
\author[3]{Xuejun Liao}
\author[3]{Lawrence Carin}
\author[4]{Jonathan P. How}

\affil[1]{Laboratory for Information and Decision Systems \\
MIT, Cambridge, MA 02139, USA\\
miaoliu@mit.edu}
\affil[2]{Department of Computer Science \\
University of New Hampshire, Durham, NH 03824, USA \\
camato@cs.unh.edu}
\affil[3]{Department of Electrical \& Computer Engineering \\
Duke University, Durham NC 27708, USA \\
\{xjliao,lcarin\}@duke.edu}
\affil[4]{Department of Aeronautics \& Astronautics \\
MIT, Cambridge, MA 02139, USA \\
jhow@mit.edu}
\date{}

\usepackage[normalem]{ulem}

\def\abovestrut#1{\rule[0in]{0in}{#1}\ignorespaces}
\def\belowstrut#1{\rule[-#1]{0in}{#1}\ignorespaces}
\def\abovespace{\abovestrut{0.20in}}

\def\belowspace{\belowstrut{0.10in}}

\usepackage{graphicx} 
\usepackage{subfigure}
\usepackage{float}
\usepackage{natbib}

\usepackage{algorithmic}

\usepackage{bm}

\usepackage{times}

\usepackage{microtype}
\usepackage{url}
\PassOptionsToPackage{hyphens}{url}
\usepackage{hyperref}
\usepackage{amssymb,amsthm,amsmath,bm,booktabs}
\usepackage{graphicx} 
\usepackage{subfigure} 
\usepackage{wrapfig}
\usepackage[font=small,labelfont=bf]{caption}
\usepackage{caption} 
\newcounter{mytempeqncnt}

\usepackage{color}
\usepackage{soul,balance}

\usepackage{comment}

\usepackage{pgf,epsf,bm,amssymb,amsthm,amsmath,array,makecell}
\usepackage{hvfloat,lipsum}
\usepackage{pgf,amssymb,amsthm,amsmath}

\newtheorem{thm}{Theorem} 

\newtheorem{definition}[thm]{Definition}

\usepackage[ruled,lined,boxed,commentsnumbered]{algorithm2e}

\makeatletter
\newcommand{\removelatexerror}{\let\@latex@error\@gobble}
\makeatother

\definecolor{orange}{rgb}{1,0.5,0}
\definecolor{purple}{rgb}{0.5,0,0.5}
\definecolor{green}{rgb}{0.1,1,0.1}
\begin{document}

\maketitle
\begin{abstract}
  Expectation maximization (EM) has recently been shown to be an efficient algorithm for learning finite-state controllers (FSCs) in large decentralized POMDPs (Dec-POMDPs). However, current methods use fixed-size FSCs and often converge to maxima that are far from optimal.  This paper considers a variable-size FSC to represent the local policy of each agent. These variable-size FSCs are constructed using a stick-breaking prior, leading to a new framework called \emph{decentralized stick-breaking policy representation} (Dec-SBPR). This approach learns the controller parameters with a variational Bayesian algorithm without having to assume that the Dec-POMDP model is available. The performance of Dec-SBPR is demonstrated on several benchmark problems, showing that the algorithm scales to large problems while outperforming other state-of-the-art methods.
\end{abstract}

\section{Introduction}
\label{introduction}
Decentralized partially observable Markov decision processes (Dec-POMDPs)~\cite{CDC13,Oliehoek:CH2012} provide a general framework for solving the cooperative multiagent sequential decision-making problems that arise in numerous applications, including robotic soccer~\cite{Messias:AAMAS10},  transportation~\cite{ICRA15MacDec}, extraplanetary exploration~\cite{Bernstein01}, and traffic control~\cite{WU:IJCAI2013}. Dec-POMDPs can be viewed as a POMDP controlled by multiple distributed agents. These agents make decisions based on their own local streams of information (i.e., observations), and their joint actions control the global state dynamics and the expected reward of the team. Because of the decentralized decision-making, an individual agent generally does not have enough information to compute the global belief state, which is a sufficient statistic for decision making in POMDPs. This makes generating an optimal solution in a Dec-POMDP more difficult than for a POMDP \cite{Bernstein:MOR-2002}, especially for long planning horizons. 

To circumvent the difficulty of solving long-horizon Dec-POMDPs optimally, while still generating a high quality policy, this paper presents scalable learning methods using a finite memory 
policy representation. 
For infinite-horizon problems (which continue for an infinite number of steps), significant progress has been made with agent policies represented as finite-state controllers (FSCs) that map observation histories to actions \cite{BernsteinJAIR09,Amato:JAAMAS-2009}.
Recent work has shown that expectation-maximization (EM)~\cite{EM77Dempster} is a scalable method for generating controllers for large Dec-POMDPs \cite{Kumar:UAI10,JJ:NIPS11}. In addition, EM has also been shown to be an efficient algorithm for policy-based reinforcement learning (RL) in Dec-POMDPs, where agents learn FSCs based on trajectories, without knowing or learning the Dec-POMDP model~\cite{WU:IJCAI2013}. 

An important and yet unanswered question is how to define an appropriate number of nodes in each FSC. 
Previous work assumes a fixed FSC size for each agent, but the number of nodes affects both the quality of the policies and the convergence rate.  When the number of nodes is too small, the FSC is unable to represent the optimal policy and therefore will quickly converge to a sub-optimal result. By contrast, when the number is too large, the FSC overfits data, often yielding slow convergence and, again, a sub-optimal policy. 

This paper uses a Bayesian nonparametric approach to determine the appropriate controller size in a variable-size FSC. 
Following previous methods \cite{WU:IJCAI2013,Oliehoek:CH2012}, learning is assumed to be centralized, and execution is decentralized. That is, learning is accomplished offline based on all available information, but the optimization is only over decentralized solutions.
Such a controller is constructed using the stick-breaking (SB) prior \cite{ishwaran2001gibbs}. 
The SB prior allows the number of nodes to be variable, but 
the set of nodes that is actively used by the controller is encouraged to be compact. 
The nodes that are actually used are determined by the posterior, combining the SB prior and the information from trajectory data. 
The framework is called the \emph{decentralized stick-breaking policy representation} (Dec-SBPR) to recognize the role of the SB prior.




In addition to the use of variable-size FSCs, the paper also makes several other contributions. Specifically, our algorithm directly operates on the (shifted) empirical value function of Dec-POMDPs, which is simpler than the likelihood functions (a mixture of dynamic Bayes nets (DBNs)) in existing planning-as-inference frameworks~\cite{Kumar:UAI10,WU:IJCAI2013}. 
Moreover, 
we derive a variational Bayesian (VB) algorithm for learning the Dec-SBPR based only on the agents' trajectories (or episodes) of actions, observations, and rewards. The VB algorithm is linear in the number of agents and at most square in the problem size, and is therefore scalable to large application domains. In practice, these trajectories can be generated by a simulator or a set of real-world experiences that are provided, and this batch data scenario is general and realistic, as it is widely adopted in learning from demonstration~\cite{martins2010learning}, and reinforcement learning. To the best of our knowledge, this is the first application of Bayesian nonparametric methods to the difficult and little-studied problem of policy-based RL in Dec-POMDPs, and the proposed method is able to generate high-quality solutions for large problems.   

\section{Background and Related Work}
Before introducing the proposed method, we first describe the Dec-POMDP model and some related work.
\subsection{Decentralized POMDPs}\label{sec:intro}
A Dec-POMDP can be represented as a tuple $\mathcal{M}=\langle \mathcal{N}, \mathcal{A},\mathcal{S}, \mathcal{O}, \mathcal{T}, \Omega, \mathcal{R}, \gamma \rangle$, where $\mathcal{N} =\{1,\cdots, N\}$ is a finite set of agent indices; $\mathcal{A}=\otimes_{n} \mathcal{A}_n$ and $\mathcal{O}=\otimes_n \mathcal{O}_n$ respectively are sets of joint actions and observations, with $\mathcal{A}_n$ and $\mathcal{O}_n$ available to agent $n$. At each step, a joint action $\vec{a}=(a_1, \cdots, a_N)\in\mathcal{A}$ is selected and a joint observation $\vec{o}=(o_1, \cdots, o_N)$ is received; $\mathcal{S}$ is a set of finite world states; $\mathcal{T} : \mathcal{S}\times \mathcal{A}\times \mathcal{S}\rightarrow [0,1]$ is the state transition function with $\mathcal{T}(s'|s,\vec{a})$ denoting the probability of transitioning to $s'$ after taking joint action $\vec{a}$ in $s$; $\Omega : \mathcal{S}\times \mathcal{A}\times\mathcal{O}\rightarrow [0,1]$ is the observation function with $\Omega(\vec{o}|s', \vec{a})$ the probability of observing $\vec{o}$ after taking joint action $\vec{a}$ and arriving in state $s'$; $\mathcal{R} : \mathcal{S}\times \mathcal{A}\rightarrow{\mathbb{R}}$ is the reward function with $r(s, \vec{a})$ the immediate reward received after taking joint action $\vec{a}$ in $s$; $\gamma\in[0,1)$ is a discount factor.
A global reward signal is generated for the team of agents after joint actions are taken, but each agent only observes its local observation. 
Because each agent lacks access to other agents' observations, each agent maintains a local policy $\pi_n$, defined as a mapping from local observation histories to actions. A joint policy consists of the local policies of all agents. For an infinite-horizon Dec-POMDP with initial belief state $b_0$, the objective is to find a joint policy $\Psi=\otimes_n \Psi_n$, such that the value of  $\Psi$ starting from $b_0$, $V^\Psi(b(s_0)) = \mathbb{E}\big[\sum_{t=0}^{\infty}\gamma^tr(s_t, \vec{a}_t)|b_0, \Psi\big]$, is maximized.

An FSC is a compact way to represent a policy as a mapping from histories to actions.
Formally, a stochastic FSC for agent $n$ is defined as a tuple $\Theta_n=\langle \mathcal{A}_n, \mathcal{O}_n, \mathcal{Z}_n,\mu_n,W_n,\pi_n \rangle$, where, $\mathcal{A}_n$ and $\mathcal{O}_n$ are the same as defined in the Dec-POMDP; $\mathcal{Z}_n$ is a finite set of controller nodes for agent $n$; $\mu_n$ is the initial node distribution with $\mu^z_{n}$ the probability of agent $n$ initially being in $z$; $W_n$ is a set of Markov transition matrices with $W^{z,z'}_{n,a,o}$ denoting the probability of the controller node transiting from $z$ to $z'$ when agent $n$ takes action $a$ in $z$ and sees observation $o$; $\pi_n$ is a set of stochastic policies with $\pi^{a}_{n, z}$ the probability of agent $n$ taking action $a$ in $z$.

{For simplicity, we use the following notational conventions. $\mathcal{Z}_n=\{1,2,\cdots,C_n\}$, where $C_n\stackrel{def.}{=}|\mathcal{Z}_n|$ is the cardinality of $\mathcal{Z}_n$, and $\mathcal{A}_n$ and $\mathcal{O}_n$ follow similarly. $\Theta=\{\Theta_{1},\cdots,\Theta_{N}\}$ is the joint FSC of all agents. A consecutively-indexed variable is abbreviated as the variable with the index range shown in the subscript or superscript; when the index range is obvious from the context, a simple ``$:$'' is used instead. Thus, $a_{n, 0:T}=(a_{n,0}, a_{n, 1}$ $,\dots,a_{n, T})$ represents the actions of agent $n$ from step $0$ to $T$ and $W^{z,:}_{n,a,o}=(W^{z,1}_{n,a,o},W^{z,2}_{n,a,o},\cdots,W^{z,|\mathcal{Z}_n|}_{n,a,o})$ represents the node transition probabilities for agent $n$ when starting in node $z$, taking action $a$ and seeing observation $o$. Given $h_{n,t}=\{a_{n,0:t-1},o_{n, 1:t}\}$, a local history of actions and observations up to step $t$, as well as an agent controller, $\Theta_n$, we can calculate a local policy $p(a_{n,t}|h_{n,t},\!\Theta_n)$, the probability that agent $n$ chooses its action $a_{n, t}$. 
		
\subsection{Planning as Inference in Dec-POMDPs}\label{sec:EM}
A Dec-POMDP planning problem can be transformed into an inference problem and then efficiently solved by EM algorithms. The validity of this method is based on the fact that by introducing binary rewards $R$ such that $P(R=1|a ,s)\propto r(a,s), \forall a\in\mathcal{A}, s\in\mathcal{S}$ and choosing the geometric time prior $p(T)=\gamma^T(1-\gamma)$, maximizing the likelihood $L(\Theta) = P(R=1;\Theta) = \sum_{T=0}^{\infty}P(T)P(R=1|T;\Theta)$ of a mixture of dynamic Bayes nets is equivalent to optimizing the associated Dec-POMDP policy, as the joint-policy value $V(\Theta)$ and $L(\Theta)$ can be related through an affine transform~\cite{Kumar:UAI10}
	\begin{eqnarray}\label{eq:likelihood}
	\label{eq:valuelikelihood}
	\hspace{-0.15cm}\mbox{$$}L(\Theta)\!=\! \frac{(1-\gamma)(V(\Theta)\!-\!\sum_T\!\gamma^TR_{min})}{R_{max}-R_{min}}  \!=\!\frac{(1-\gamma)\hat{V}(\Theta)}{R_{max}-R_{min}},\!\!\!\!\!
	\end{eqnarray}
where $R_{max}$ and $R_{min}$ are maximum and minimum reward values and $\!\hat{V}(\Theta)\!\!\stackrel{def.}{=}\!\!\!V(\Theta)\!-\!\sum_T\!\gamma^T\!R_{min}$ is a shifted value.	

Previous EM methods~\cite{Kumar:UAI10,JJ:NIPS11} have achieved success in scaling to larger problems by factoring the distribution over states and histories for inference, but 
 these methods require using a Dec-POMDP model to construct a Bayes net for policy evaluation. When the exact model parameters $\mathcal{T}$, $\Omega$ and $\mathcal{R}$ are unknown, 
 one needs to solve a reinforcement learning (RL) problem. To address this important yet less addressed problem, a global empirical value function extended from the single-agent case \cite{Li-Liao-Carin:RPR-09}, is constructed based on all the action, observation and reward trajectories, and the product of local policies of all agents. This serves as the basis for learning (fixed-size) FSCs in RL settings. 
\begin{definition} (Global empirical value function) \label{def:obj-value}Let
	$\mathcal{D}^{(K)}=\{(\vec{o}^{\,k}_{0}\vec{a}^{\,k}_{0}r^k_{0}\vec{o}^{\,k}_{1}\vec{a}^{\,k}_{1}r^k_{1}\cdots{}\vec{o}^{\,k}_{T_k}\vec{a}^{\,k}_{T_k}$ $r^k_{ T_k})\}_{k=1,\cdots, K}$
	be a set of episodes resulting from  $N$ agents who choose actions according to $\Psi=\otimes_n\Psi_n$,
	a set of stochastic behavior policies with $p^{\Psi_{n}}(a|h)>0$, $\forall$ action $a$, $\forall$
	history $h$. The global empirical value function is defined as
	$\hat{V}\big(\mathcal{D}^{(K)};\Theta\big)\!\stackrel{def.}{=}\!
	\sum_{k=1}^K\sum_{t=0}^{T_k}\frac{\gamma^t(r_t^k-R_{min})}{K}
	\frac{\prod_{\tau=0}^t\prod_{n=1}^N p(a^k_{n, \tau}|h^k_{n, \tau},\Theta_n)}{\prod_{\tau=0}^t\prod_{n=1}^N p^{\Psi_n}(a^k_{n, \tau}|h^k_{n, \tau})}$
	where $h_{n,t}^{\,k}=(a^k_{n,0:t-1},o^k_{n,1:t})$, $0\leq\gamma<1$ is the discount.
\end{definition}

According to the strong law of large numbers~\cite{robert2004monte}, $\hat{V}(\Theta)=\lim_{K \!\rightarrow\infty}\hat{V}\big(\mathcal{D}^{(K)};\Theta\big)$, i.e., with a large number of trajectories, the empirical value function $\hat{V}\big(\mathcal{D}^{(K)};\Theta\big)$ approximates $\hat{V}(\Theta)$ accurately. Hence, applying~\eqref{eq:valuelikelihood}, $\hat{V}\big(\mathcal{D}^{(K)};\Theta\big)$ approximates $L(\Theta)$, and offers an objective for learning the decentralized policies and can be directly maximized by the EM algorithms in \cite{Li-Liao-Carin:RPR-09}. 

\section{Bayesian Learning of Policies}
EM algorithms infer policies based on fixed-size representation and observed data only, it is difficult to explicitly handle model uncertainty and encode prior (or expert) knowledge. To address these issues, a Bayesian learning method is proposed in this section. This is accomplished by measuring the likelihood of $\Theta$ using $L\big(\mathcal{D}^{(K)};\Theta\big)$, which is combined with the prior $p(\Theta)$ in Bayes' rule to yield the posterior
\begin{eqnarray}
p(\Theta|\mathcal{D}^{(K)}) = L\big(\mathcal{D}^{(K)};\Theta\big) p(\Theta)\left[p\big(\mathcal{D}^{(K)}\big)\right]^{-1},
\end{eqnarray}
where $p\big(\mathcal{D}^{(K)}\big)$ is the marginal likelihood of the joint FSC and, up to additive constant, proportional to the marginal value function,
	\begin{eqnarray}
	&&\hspace{-2.0cm}\hat{V}\big(\mathcal{D}^{(K)}\big)\!\stackrel{def.}{=}\!\int \hat{V}\big(\mathcal{D}^{(K)};\Theta\big) p(\Theta)d\Theta
	\cr
	&&\hspace{-0.5cm}\mbox{$\propto\!\int L\big(\mathcal{D}^{(K)};\Theta\big) p(\Theta)d\Theta = p\big(\mathcal{D}^{(K)}\big)$}.
	\end{eqnarray}
To compute the posterior, $p(\Theta|\mathcal{D}^{(K)})$, Markov chain Monte Carlo (MCMC) simulation~\cite{robert2004monte} is the most straight forward method. However, MCMC is costly in terms of computation and storage, and lacks a strong convergence guarantee. An alternative is a variational Bayes (VB) method~\cite{beal2003variational}, which performs approximate posterior inference by minimizing the Kullback-Leibler (KL) divergence between the true and approximate posterior distributions. Because the VB method has a (local) convergence guarantee and is able to trade-off scalability and accuracy, we focus on the derivation of VB method here. Denoting $q(\Theta)$ as the variational approximation to $p(\Theta|\mathcal{D}^{(K)})$, and $q_t^k(\vec{z}_{0:t}^{\,k})$ as the approximation to $p(\vec{z}_{0:t}^{\,k}|\vec{o}_{1:t}^{\,k},\Theta)$, a VB objective function~\footnote{Refer to the appendix for derivation details} is

\begin{eqnarray}\label{eq:KL}
&&\hspace{-1cm}\mathrm{KL}\big(\big\{q_t^k(\vec{z}_{0:t}^{\,k})q(\Theta)\big\}_{k=1:K}||\big\{\nu^{\,k}_tp(\vec{z}_{0:t}^{\,k},\Theta)\big\}_{k=1:K}\big)={ln}\hat{V}\big(\mathcal{D}^{(K)}\big)-\mathrm{LB}\big(\big\{q_t^k(\vec{z}_{0:t}^{\,k})\big\}, q(\Theta)\big),
\end{eqnarray}
where
\begin{eqnarray}\label{eq:vblb}
&&\hspace{-1.0cm}\mathrm{LB}\big(\big\{q_t^k(\vec{z}_{0:t}^{\,k})\big\}, q(\Theta)\big)\stackrel{def.}{=}\sum_{k,t,z_{1:N,0:t}^{\,k}} \int \frac{q_t^k(\vec{z}_{0:t}^{\,k})}{K/q(\Theta)}\ln\frac{\nu_t^kp(\vec{z}_{0:t}^k,\Theta|h_{0:t}^k)}{q_t^k(\vec{z}_{0:t}^{\,k})q(\Theta)}d\Theta
\end{eqnarray}
is the lower bound of $\mathrm{ln}\hat{V}\big(\mathcal{D}^{(K)}\big)$ and
\begin{eqnarray}\label{eq:reweightedreward}
			\hspace{-0.5cm}\nu^{\,k}_t&\!\!\!\stackrel{def.}{=}\!\!\!&\frac{\gamma^tr^k_t \prod_{n=1}^N p(a^k_{n, 0:t}|o^k_{n, 1:t})}{\prod_{n=1}^N\prod_{\tau=0}^tp^{\Psi_n}(a^k_{n, \tau}|h^k_{n, \tau})\widehat{V}(\mathcal{D}^{(K)})}, \forall\,t,k,
\end{eqnarray}
is the re-weighted reward.
Since $\mathrm{ln}\hat{V}\big(\mathcal{D}^{(K)}\big)$ in equation~\eqref{eq:KL} is independent of $\Theta$ and $\{q_t^k(\vec{z}_{0:t}^{\,k})\}$, minimizing the KL divergence is equivalent to maximizing the lower bound, leading to the following constrained optimization problem,
\begin{eqnarray}\label{eq:VBproblem}
&&\hspace{-0.5cm}\mbox{$\max_{\big\{q_t^k\big(\vec{z}_{0:t}^{\,k}\big)\big\}q(\Theta)}  \;\; \mathrm{LB}\big(\big\{q_t^k(\vec{z}_{0:t}^{\,k})\big\}, q(\Theta)\big)$}
\cr
&&\hspace{-0.5cm}
\mbox{$\textrm{subject to:}\;\;q_t^k(\vec{z}_{0:t}^{\,k}, \Theta) 
	= \prod_{n=1}^N q_t^k(z_{n, 0:t}^{\,k})q(\Theta_n),$}  
\cr
&&\hspace{-0.5cm}\sum_{k=1}^K\sum_{t=0}^{T_k}\sum_{z_{1:N,0:t}^{\,k}=1}^{|\mathcal{Z}|}\!\!\!\!\!\!\!\! q_t^k(\vec{z}_{0:t}^{\,k}) = K,\,\,\,\,q_t^k(\vec{z}_{0:t}^{\,k}) \geq 0, \forall \vec{z}_t^{\,k}, t, k,
\cr
&&\hspace{-0.1cm}\mbox{$\int p(\Theta)d\Theta = 1 \;\;\text{and}\;\; p(\Theta)\geq 0, \forall \Theta, $}
\end{eqnarray}
where the constraint in the second line arises both from the mean-field approximation and from the decentralized policy representation, and the last two lines summarize the normalization constraints. It is worth emphasizing that we developed this variational mean-field approximation to optimize a decentralized policy representation, showing that the VB learning problem formulation~\eqref{eq:VBproblem} is both a general and accurate method for the multiagent problem considered in this paper.

\subsection{Stick-breaking Policy Priors}\label{sec:SBprior}
To solve the Bayesian learning problem described above and obtain the variable-size FSCs, the stick-breaking prior is used to specify the policy's structure. As such, Dec-SBPR is formally given in definition~\ref{def:Dec-SBPR}.
\begin{definition}\label{def:Dec-SBPR}
	The decentralized stick breaking policy representation (Dec-SBPR) is a tuple 
	($\mathcal{N}, \mathcal{A}, \mathcal{O},$ $ \mathcal{Z}, {\mu}, {\eta},{\rho}$), where
	$\mathcal{N}, \mathcal{A}$ and $\mathcal{O}$ are as in the
        definition of Dec-POMDP; $\mathcal{Z}$ is an \emph{unbounded}
        set of nodes indexed by positive integers; for notational
        simplicity\footnote{Nonparametric priors over $\mu$ can also be used.},${\mu}$ are assumed to be deterministic with $\mu_n^1=1,\mu_n^{2:\infty}=0, \forall n$; $({\eta},{\rho})$ determine $(W,\pi)$, the FSC parameters defined in section~\ref{sec:intro}, as follows
	\begin{eqnarray}\label{eq:Dec-SBPR}
	&&\hspace{-1cm}W_{n,a,o}^{i, 1:\infty}{\sim}\mathrm{SB}({\sigma}_{n,a,o}^{i, 1:\infty}, {\eta}_{n,a,o}^{i, 1:\infty}),\;\;\pi^{1:|\mathcal{A}_n|}_{n,i}\sim \mathrm{Dir}(\rho^{1:|\mathcal{A}_n|}_{n,i})
	\end{eqnarray}
	where $\mathrm{Dir}$ represents Dirichlet distribution and $\mathrm{SB}$ represents the stick-breaking process with $W_{n,a,o}^{i,j}=V^{i,j}_{n,a,o}\prod^{j-1}_{m=1}(1-V^{i, m}_{n,a,o})$ and $V^{i,j}_{n,a,o}\sim\mathrm{Beta}({\sigma}_{n,a,o}^{i,j}, {\eta}_{n,a,o}^{i,j})$, ${\eta}_{n,a,o}^{i,j}\sim\mathrm{Gamma}(c,d), n = 1,\cdots, N$ and $i, j = 1,\cdots, \infty$.
\end{definition}

DECSBPR differs from previous nonparametric Bayesian RL methods~\cite{Liu:ICML11,doshi2010nonparametric}. Specifically, Dec-SBPR performs policy-based RL and generalizes the nonparametric Bayesian policy representation of POMDPs \cite{Liu:ICML11} to the decentralized domain. 
Whereas~\cite{doshi2010nonparametric} is a model-based RL method that doesn't assume knowledge about the world's model, but explicitly learns it and then performs planning.} Moreover, Dec-SBPR further distinguishes from previous methods~\cite{doshi2010nonparametric,Liu:ICML11} by the prior distributions and inference methods employed. These previous methods employed hierarchical Dirichlet processes hidden Markov models (HDP-HMM) to infer the number of controller nodes. However, due to the lack of conjugacy between two levels of DPs in the HDP-HMM, a \emph{fully conjugate Bayesian} variational inference does not exist\footnote{The VB method in~\cite{bryant2012truly} imposes point-mass proposals over top level DPs, lacking a uncertainty measure.}. Therefore, these methods used MCMC which requires high computational and storage costs, making them not ideal for solving large problems. In contrast, Dec-SBPR employs single layer SB priors over FSC transition matrices $W$ and sparse Gamma priors over SB weight hyperparameters $\eta$ to bias transition among nodes with smaller indices. A similar framework has been explored to infer HMMs, and we refer readers to~\cite{paisley2009hidden} for more details.
 

It is worth noting that SB processes subsume Dirichlet Processes~(DPs)~\cite{Ferguson73} as a special case, when ${\sigma}_{n,a,o}^{i,j}\!\!=\!1,\forall i,j,n,a,o$ (in Dec-SBPR). 
 The purpose of using SB priors is to encourage a small number of FSC nodes. Compared to a DP, the SB priors can represent richer patterns of sparse transition between the nodes of an FSC, because it allows arbitrary correlation between the stick-breaking weights (the weights are always negatively correlated in a DP).
\begin{algorithm}[t]		
	\begin{small}
		\caption{Batch VB Inference for Dec-SBPR}
		\label{alg:Dec-SBPR-off-policy}
		\begin{algorithmic}[1]
			\footnotesize
			\STATE {\bfseries Input:} Episodes $\mathcal{D}^{(K)}$, the number of agents $N$, initial policies $\Theta$, VB lower bound $\rm{LB}_0=-Inf$, $\Delta\rm{LB}=1$, $\rm{Iter}=0$;  
			\WHILE{$\Delta\rm{LB}>10^{-3}$}
			\FOR {$k = 1$ to $K$, $n = 1$ to $N$}
			\STATE Update the global rewards  $\{\hat{\nu}_t^k\}$ using \eqref{eq:recomputed-reward}.
			\STATE Compute $\{\alpha_{\tau}^{n,k}\}$ and $\{\beta_{t,\tau}^{n,k}\}$. 
			\ENDFOR
			\STATE $\rm{Iter} = \rm{Iter} + 1$.
			\STATE Compute $\rm{LB}_{\rm{Iter}}$ using \eqref{eq:vblb}
			\STATE $\Delta\rm{LB} = (LB_{Iter} - LB_{Iter-1})/|LB_{Iter-1}|$
			\FOR{$n = 1$ to $N$}
			\STATE Compute $\{\xi^{n, k}_{t,\tau}(i,j)\}$ and $\{\phi^{n, k}_{t,\tau}(i)\}$ using \eqref{eq:p(z1,z2|a,o)}.
			\STATE Update the hyper-parameters of $\Theta_n$ using (\ref{eq:theta-update-dec}).
			\STATE Compute $|\mathcal{Z}_n|$ using \eqref{eq:zsize}.
			\ENDFOR
			\ENDWHILE
			\STATE  {\bfseries Return:} Policies $\{\Theta_n\}_{n=1}^{N}$, and controller sizes $\{|\mathcal{Z}_n|\}_{n=1}^{N}$.
		\end{algorithmic}
	\end{small}
\end{algorithm}
\subsection{Variational Stick-breaking Policy Inference}\label{sec:vbsb_inference}
It is shown in \cite{ishwaran2001gibbs} that the random weights constructed by the SB prior are equivalently governed by a generalized Dirichlet distribution (GDD) and are therefore conjugate to the multinomial distribution; hence an efficient variational Bayesian algorithm for learning the decentralized policies can be derived. 
To accommodate an unbounded number of nodes, we apply the retrospective representation of SB priors  \cite{papaspiliopoulos2008retrospective} to the Dec-SBPR. For agent $n$, the SB prior is set with a truncation level $|\mathcal{Z}_n|$, taking into account the current occupancy as well as additional nodes reserved for future new occupancies. 
The solution to (\ref{eq:VBproblem}) under the stick-breaking priors is given in Theorem \ref{theorem:RPR-VB}, the proof of which is provided in the appendix.
\begin{thm}
	\label{theorem:RPR-VB} Let $p(\Theta)$ be constructed by the SB priors defined in~\eqref{eq:Dec-SBPR} with hyper-parameters
	$(\hat{\sigma},\hat{\eta},\hat{\rho})$, then iterative application of the following
	updates leads to monotonic increase of  (\ref{eq:vblb}), until convergence to a maxima. The updates of $\{q^k_t\}$ are 
	\begin{eqnarray} \label{eq:recomputed-reward}
		\mbox{$q^k_t(z^k_{n, 0:t})=\hat{\nu}_t^kp(z^k_{n, 0:t}|o^k_{n, 1:t}, a^k_{n, 0:t}, \widetilde{\Theta}_n), \forall\,n, t,k,$}
	\end{eqnarray}
	where $\hat{\nu}_t^k$ is computed using \eqref{eq:reweightedreward} with $\Theta$ replaced by $\widetilde{\Theta}=\{\widetilde{\pi},\widetilde{\mu},\widetilde{W}\}$, a set of under-normalized probability (mass) functions 
	, with
	$\widetilde{\pi}^{a}_{n,i}=e^{\langle\ln \pi_{n,i}^{a}\rangle_{p(\pi|\bm{\hat{\rho}})}}$,
	and
	$\widetilde{W}^{i,j}_{n,a,o}=e^{\langle\ln W_{n,a,o}^{i, j}\rangle_{p(W|\bm{\hat{\sigma},\hat{\eta}})}}$,
	and $\langle\cdot\rangle_p$ denotes expectations of $\cdot$ with respect to distributions $p$. The hyper-parameters of the posterior distribution are updated as
	\begin{eqnarray}\label{eq:theta-update-dec}
	\label{eq:hyper-w-new1}
	&&\hspace{-0.7cm}\hat{{\sigma}}^{i, j}_{n,a,o}={\sigma}^{i, j}_{n,a,o}+{\zeta}^{i,j}_{n,a,o},\;\; \hat{{\eta}}^{i, j}_{n,a,o} = {\eta}^{i, j}_{n,a,o}+ \sum_{l=j+1}^{|\mathcal{Z}_n|}{\zeta}^{i,l}_{n,a,o},
	\cr
	&&\hspace{-0.7cm}\hat{\rho}_{n, i}^{a}=\rho^a_{n, i} +\sum_{k=1}^K\sum_{t=0}^{T_k}\sum_{\tau=1}^{t}\frac{\hat{\nu}^{\,k}_t}{K}\phi_{t,\tau-1}^{n, k}(i)\mathbb{I}_a(a_{\tau}^{\,k})
	\end{eqnarray}
	with ${\zeta}^{i,j}_{n,a,o}\!\!=\!\!\sum_{k=1}^K\sum_{t=0}^{T_k}\sum_{\tau=1}^{t}\frac{\hat{\nu}^{\,k}_t}{K}\!
	\xi_{t,\tau-1}^{n, k}(i,j)\mathbb{I}_{a,o}(a_{\tau-1}^{\,k},o_{\tau}^{\,k})$, where $\mathbb{I}(\cdot)$ is the indicator function, and both $\xi^{n, k}_{t,\tau}$ and $\phi^{n, k}_{t,\tau}$ are marginals of $q^k_t(z^k_{n, 0:t})$, i.e.
\begin{eqnarray}\label{eq:p(z1,z2|a,o)}
&&\hspace{-1cm}\xi^{n, k}_{t,\tau}(i, j)\!=\!p(z^k_{n, \tau}=i,z^k_{n, \tau+1}=j|a^k_{n, 0:t}, o^k_{n, 1:t},\widetilde{\Theta}_n)
\\\label{eq:p(z|a,o)}
&&\hspace{-1cm}\phi^{n,k}_{t,\tau}(i)\!=\!p(z^k_{n, \tau}=i|a^k_{n, 0:t}, o^k_{n, 1:t},\widetilde{\Theta}_n)
\end{eqnarray}
\end{thm}
The update equations in Theorem \ref{theorem:RPR-VB} constitute the VB algorithm for learning a variable-size joint FSCs under SB priors with batch data. In particular, \eqref{eq:recomputed-reward} is a policy-evaluation step where the rewards are reweighted to reflect the improved marginal value of the new policy posterior updated in the previous iteration, and \eqref{eq:theta-update-dec} is a policy-improvement step where the reweighted rewards are used to further improve the policy posterior. Both steps require (\ref{eq:p(z1,z2|a,o)}), which are computed based on $\alpha_{\tau}^{n, k}(i)\!\!=\!\!p\big(z_{n, \tau}^{k}\!\!=\!\!i | a_{n, 0:\tau}^{\,k}, o_{n, 1:\tau}^{\,k}, \widetilde{\Theta}_n\big)$ and $ \beta_{t,\tau}^{n, k}(i)\!\!=\!\!\frac{p(a_{n, \tau+1:t}^{\,k} | z_{n, \tau}^{\,k}=i, o_{n, \tau+1:t}^{\,k}, \widetilde{\Theta}_n)}{\prod_{\tau'=\tau}^t p(a_{\tau}^{\,k}| h_{n, \tau'}^{\,k}, \widetilde{\Theta}_n)}$, $\forall\,n,k,t,\tau$.  The $(\alpha,\beta)$ are forward-backward messages. Their updating equations are derived in the appendix.

To determine the number of controller nodes $\{|\mathcal{Z}_n|\}_{n=1}^N$, the occupancy of a node is computed by checking if there is a positive reward assigned to it. For example, for action $a$ and node $i$, $\hat{\rho}^{a}_{n, i}-\rho^{a}_{n, i}$ is the reward being assigned. If this quantity is greater than zero, then node $i$ is visited. Summing over all actions gives the value of node $i$. Hence $|\mathcal{Z}_n|$ can be computed based on the following formula
\begin{eqnarray}\label{eq:zsize}
	|\mathcal{Z}_n| = \mbox{$\sum_{i=1}^\infty \mathbb{I}\big(\sum_{a=1}^{|\mathcal{A}_n|}$}(\hat{\rho}_{n, i}^{a}-\rho^a_{n, i})>0\big).
\end{eqnarray}
The complete algorithm is described in Algorithm \ref{alg:Dec-SBPR-off-policy}. Upon the convergence of Algorithm \ref{alg:Dec-SBPR-off-policy}, point estimates of the decentralized policies may be obtained by calculating the expectation: $\mathbb{E}[\hat{\mu}^i_n]$, $\mathbb{E}[\hat{\pi}_{n, i}^{a}]$, and $\mathbb{E}[\hat{W}^{i, j}_{n,a,o}]$ (see the appendix for details).

\begin{table}[h]
	\caption{Computational Complexity of Algorithm \ref{alg:Dec-SBPR-off-policy}.}
	\label{tab:complexity}
	\begin{center}
		\begin{small}
			\begin{sc}
				\begin{tabular}{lccr}
					\hline				
					var & best case & worst case \\
					\hline 
					$\alpha$ & $\Omega(N|\mathcal{Z}|^2KT)$ & $O(N|\mathcal{Z}|^2KT)$  \\ 
					$\beta$ & $\Omega(N|\mathcal{Z}|^2KT)$ & $O(N|\mathcal{Z}|^2KT^2) $ \\ 
					$\nu_t^k$ & $\Omega(K)$ & $O(KT)$         \\
					$\Theta$  & $\Omega(N|\mathcal{Z}|^2KT)$ &$O(N|\mathcal{Z}|^2KT^2)$\\
					$\mathrm{LB}$ & $\Omega(|\mathcal{Z}|^2\sum_{n=1}^N|\mathcal{A}_n||\mathcal{O}_n|)$ & $O(|\mathcal{Z}|^2\sum_{n=1}^N|\mathcal{A}_n||\mathcal{O}_n|)$\\
					\hline
				\end{tabular}
			\end{sc}
		\end{small}
	\end{center}
\end{table}
\subsection{Computational complexity}
The time complexity of Algorithm \ref{alg:Dec-SBPR-off-policy} for each iteration is summarized in Table \ref{tab:complexity}, assuming the length of an episode is on the order of magnitude of $T$, and the number of nodes per controller is on the order of magnitude of $|\mathcal{\mathcal{Z}}|$. In Table \ref{tab:complexity}, the worst case refers to when there is a nonzero reward at every time step of an episode (dense rewards), while the best case is when nonzero reward is received only at the terminal step. Hence in general, the algorithm scales linearly with the number of episodes and the number of agents. The time dependency on $T$ is between linear and quadratic. In any case, the computational complexity of Algorithm~\ref{alg:Dec-SBPR-off-policy} is independent of the number of states, making it is scalable to large problems.

\subsection{Exploration and Exploitation Tradeoff\label{sec:exploration}}\label{sec:explore-exploit} 
Algorithm \ref{alg:Dec-SBPR-off-policy} assumes off-policy batch learning where trajectories are collected using a separate behavior policy. 
This is appropriate when data has been generated from real-world or simulated experiences without any input from the learning algorithm (e.g., learning from demonstration).  
Off-policy learning is efficient if the behavior policy is close to optimal, as in the case when expert information is available to guide the agents. With a random behavior policy, it may take a long time for the policy to converge to optimality; in this case, the agents may want to exploit the policies learned so far to speed up the learning process.  

An important issue concerns keeping a proper balance between exploration and exploitation
 to prevent premature convergence to a suboptimal policy, but allow the algorithm to learn quickly.
 Since the execution of Dec-POMDP policies is decentralized, it is difficult to design an efficient exploration strategy that guarantees optimality. 
 \cite{WU:IJCAI2013} count the visiting frequency of FSC nodes and apply upper-confidence-bound style heuristic to select next controller nodes, and use $\epsilon$-greedy strategy to select actions. However $\epsilon$-greedy might be sample inefficient. \cite{Banerjee:AAAI2012} proposed a distributed learning approach where agents take turns to learn the best response to each other's policies. This framework applies an R-max type of heuristic, using the counts of trajectories to distinguish known and unknown histories, to tradeoff exploration and exploitation. However, this method is confined to tree-based policies in finite-horizon problems, and requires synchronized multi-agent learning. 

To better accommodate our Bayesian policy learning framework for RL in infinite-horizon Dec-POMDPs, we define an auxiliary FSC, $\Omega_n=\langle \mathcal{Y}, \mathcal{O}_n, \mathcal{Z}_n, W_n,\mu_n,\varphi_n \rangle$, to represent the policy of each agent in balancing exploration and exploitation. To avoid confusion, we refer to $\Theta_n$ as a primary FSC. The only two components distinguishing $\Psi_n$ from  $\Theta_n$ are $\mathcal{Y}$ and $\varphi_n$, where $\mathcal{Y}=\{0,1\}$ encodes exploration ($y=1$) or exploitation ($y=0$), and  $\varphi_n=\{\varphi_y^{n,z}\}$ with $\varphi_y^{n,z}$ denoting the probability of agent $n$ choosing $y$ in $z$. One can express $p(y_{n,t}|h_{n,t},\Psi_n)$ in the same way as one expresses $p(a_{n,t}|h_{n,t},\Theta_n)$ (which is described in section~\ref{sec:intro}).
The behavior policy $\Pi_n$ of agent $n$ is given as 
\begin{eqnarray}\label{eq:behaviorpolicy}
p^{\Pi_n}(a|h,\Theta_n,\Omega_n)=\sum_{y=0,1}p(a|y,h)p(y|h,\Omega_n),
\end{eqnarray}
where $p(a|y=0,h)\equiv{}p(a|h,\Theta_n)$ is the primary FSC policy, and $p(a|y=1,h)$ is the exploration policy of agent $n$, which is usually a uniform distribution. 

The behavior policy in (\ref{eq:behaviorpolicy}) has achieved significant success in the single-agent case \cite{Cai:NIPS09,Liu:ICML11,liu2013online}. Here we extend it to the multi-agent case (centralized learning and decentralized exploration/execution) and provide empirical evaluation in the next section. 

\section{Experiments}\label{sec:experiments}
The performance of the proposed algorithms are evaluated on five
benchmark problems~\cite{URL:benchmarks} and a
large-scale problem (traffic control) \cite{WU:IJCAI2013}.
The experimental procedure in \cite{WU:IJCAI2013} was used for all the
results reported here. For Dec-SBPR, the hyperparameters in
(\ref{eq:Dec-SBPR}) are set to $c=0.1$ and $d=10^{-6}$ to promote sparse
usage of FSC nodes.\footnote{These values were chosen for testing, but our approach is robust to other values of $c$ and $d$.}
The policies are initialized as FSCs converted from
the episodes with the highest rewards using a method similar
to~\cite{CA:AAMAS09}. 
\begin{figure}[!t]
	\begin{center}
		$\begin{array}{cc}
		\hspace{-0.25cm}\includegraphics[width=2.5in]{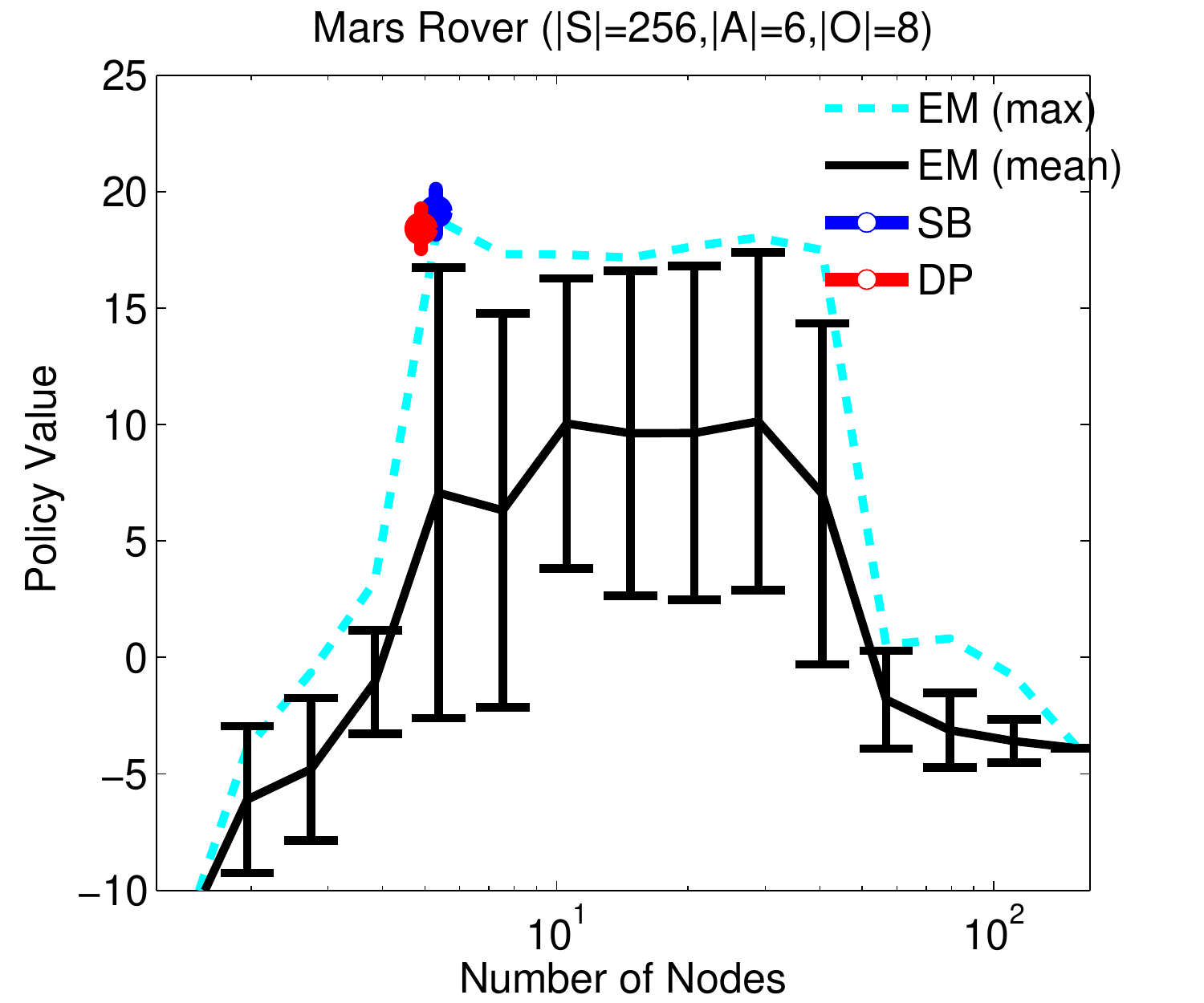}
		&\hspace{-0.0cm}\includegraphics[width=2.5in]{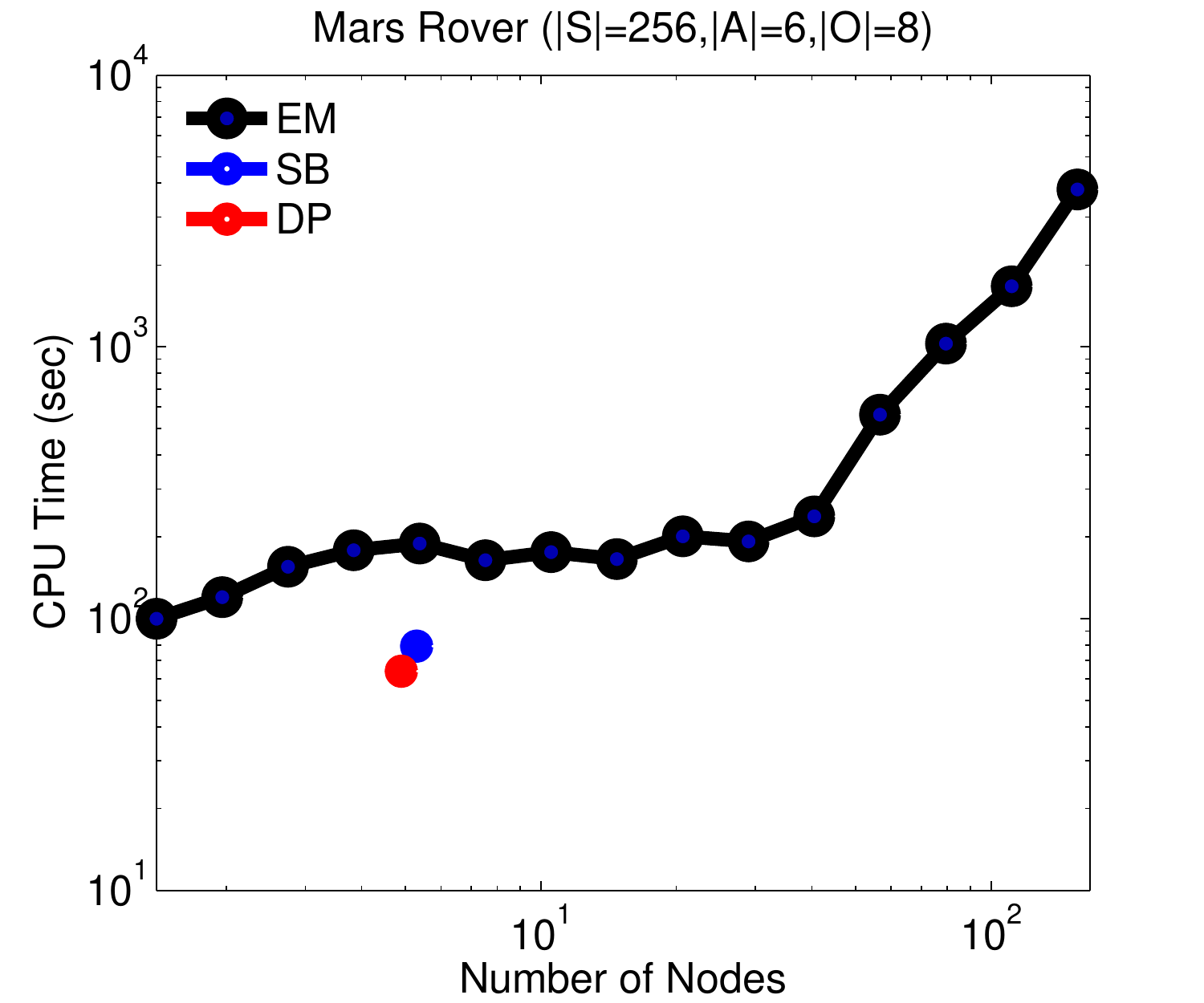}
		\end{array}$
		
		\caption{\label{fig:resultsZ} Comparison between the variable-size controller learned by Dec-SBPR and fixed-size controllers learned by  EM. Left: testing value; Right: averaged computation time. Although EM has less computational complexity than our VB algorithm per iteration, empirically, the VB algorithm uses less time to reach convergence. Moreover, the average value of using the SB prior and its special Dirichlet instance (DP) are close to the best result of EM  (dotted sky-blue line), but are much better than the average results of EM (solid back line). Using the SB prior achieves slightly better performance than the DP, which can be explained the flexibility of SB prior, as explained by the last paragraph of section~\ref{sec:SBprior}.}
	\end{center}
\end{figure}
\begin{table*}[t]
\caption{Performance of Dec-SBPR on benchmark problems compared to other
  state-of-art algorithms. Shows policy values
  (higher value indicates better performance) and CPU times of all
  algorithms, and the average controller size $|Z|$ inferred by Dec-SBPR.} 
	\label{tab:compare}
	\begin{center}
		\begin{footnotesize}
			\begin{sc}
				\resizebox{16.5cm}{!}{
					\begin{tabular}{l c ccccccc cr}
						\toprule
						\multicolumn{4}{c}{\;\;\;\;\;\;\;\;\;\;\;\;\;\;\;\;\;\;\;\;\;\;\;\;\;\;\;\;\;\;\;\;\;\;\;\;\;\;\;policy learning (unknown model)} &\multicolumn{1}{c}{\;\;\;\;\;\;\;\;\;\;\;\;\;\;\;\;\;\;\;\;\;\;\;\;\;\;\;\; planning (known model)} \\
						\cmidrule(r){2-4}\cmidrule(r){5-6}
						Problems $(|\mathcal{S}|,|\mathcal{A}|,|\mathcal{O}|)$ & Dec-SBPR(fixed iteration) & Dec-SBPR(fixed time) & MCEM &  PeriEM & FB-HSVI\\
						& 
						\begin{tabular}{c*{3}{c}}
							Value & $|Z|$ & Time
						\end{tabular}
						&
						\begin{tabular}{c*{3}{c}}
						Value & $|Z|$ & Time
						\end{tabular}
						&
						\begin{tabular}{c*{3}{c}}
							Value & $|Z|$ & Time
						\end{tabular} 
						& 
						\begin{tabular}{c*{3}{c}}
							Value & $|Z|$ & Time
						\end{tabular}
						& 
						\begin{tabular}{c*{3}{c}}
							Value & $|Z|$ & Time
						\end{tabular}
						\\
						\hline
						\abovespace
						Dec-Tiger (2, 3, 3) 
						&
						\begin{tabular}{c*{3}{c}}
						-18.63 & 6 & \;\;96s
						\end{tabular} 
						& 
						\begin{tabular}{c*{3}{c}}
						\;\;\;\;\;-19.42 & 8 & 20s
						\end{tabular} 
						&
						\begin{tabular}{c*{3}{c}}
						-32.31 & 3 & 20s
						\end{tabular} 
						&
						\begin{tabular}{c*{3}{c}}
						\;\;\;\;9.42 & $7\times10$ & 6540s
						\end{tabular}
						&
						\begin{tabular}{c*{3}{c}}
						\;\;\;\;13.45 & \;52 & \;\;6.0s
						\end{tabular}
						\\
						Broadcast (4, 2, 5)   
						& 
						\begin{tabular}{c*{3}{c}}
						\;9.20 & 2 & \;\;7s 
						\end{tabular}
						& 
						\begin{tabular}{c*{3}{c}}
						\;\;\;\;\;\;\;9.27 & \;\;2 & 24s 
						\end{tabular}
						& 
						\begin{tabular}{c*{3}{c}}
						\;\;9.15 & \;\;3 & \;\;24s
						\end{tabular}
						&
						\begin{tabular}{c*{3}{c}}
							--
						\end{tabular} 
						&
						\begin{tabular}{c*{3}{c}}
						\;\;\;\;\;\;\;9.27 & 102 & 19.8s 
						\end{tabular}
						\\
						Recycling Robots (3, 3, 2)  
						& 
						\begin{tabular}{c*{3}{c}}
						\;\;\;31.26 & 3 & \;\;147s
						\end{tabular}
						&
						\begin{tabular}{c*{3}{c}}
						\;\;\;\;\;\;\;25.16 & 2 & 19s
						\end{tabular}
						&
						\begin{tabular}{c*{3}{c}}
						30.78 & \;3 & 19s
						\end{tabular}
						& 
						\begin{tabular}{c*{3}{c}}
						31.80 & $6\times10$ & 272s
						\end{tabular}	
						&
						\begin{tabular}{c*{3}{c}}
						\;31.93 & 108 & 0s
						\end{tabular}
						\\
						Box Pushing (100, 4, 5)    
						& 
						\begin{tabular}{c*{3}{c}}
						\;\;77.65 & 14 & 290s
						\end{tabular}
						&
						\begin{tabular}{c*{3}{c}}
							\;\;\;\;\;\;\;58.27 & 9 & 32s
						\end{tabular}
						&
						\begin{tabular}{c*{3}{c}}
						59.95& \;3 & 32s
						\end{tabular}
					    & 
					    \begin{tabular}{c*{3}{c}}
					    106.68 & $4\times 10$ & 7164s
					    \end{tabular}
					    &
					    \begin{tabular}{c*{3}{c}}
					    \;\;\;\;\;\;\;\;224.43 & 331 & 1715.1s
					    \end{tabular}
					    \\
						\belowspace
						Mars Rovers (256, 6, 8)    
						& 
						\begin{tabular}{c*{3}{c}}
						\;\;\;\;20.62 & 5 & 1286s
						\end{tabular}
						&
						\begin{tabular}{c*{3}{c}}
							\;\;\;\;\;\;\;\;\;\;\;15.2 & 6 & 160s
						\end{tabular}
						&
						\begin{tabular}{c*{3}{c}}
						\;\;\;8.16& \;\;3 & 160s
						\end{tabular}
						&
						\begin{tabular}{c*{3}{c}}
						\;\;18.13 & $3\times 10$& 7132s
						\end{tabular}
						&
						\begin{tabular}{c*{3}{c}}
						\;\;\;\;\;\;\;\;26.94 & 136 & 74.31s 
						\end{tabular}
						\\
						\toprule
					\end{tabular}
				}
			\end{sc}
		\end{footnotesize}
	\end{center}
\end{table*}
\vspace{-0.2in}
\paragraph{Learning variable-size FSC vs learning fixed-size FSC}
To demonstrate the advantage of learning variable-size FSCs, Dec-SBPR is
compared with an implementation of the previous EM algorithm 
\cite{WU:IJCAI2013}. The comparison is for the Mars Rover problem using
$K=300$ episodes \footnote{Using smaller training sample size $K$, our method can still perform robustly, as it is shown in the appendix.} to learn the FSCs and evaluating the policy by the
discounted accumulated reward averaged over 100 test episodes of 1000 steps. Here, we consider off-policy learning and apply a semi-random
policy to collect samples. Specifically, the learning agent is allowed
access to episodes collected by taking actions according to a POMDP
algorithm (point-based value iteration (PBVI)~\cite{Pineau03}). Let
$\epsilon$ be the probability that the agents follow the PBVI policy and
$1-\epsilon$ be the probability that the agents take random actions. 
This procedure mimics the approach used in previous work \cite{WU:IJCAI2013}.
The results with $\eta=0.3$ are reported in Figure~\ref{fig:resultsZ}, which shows
the exact value and computation time as a function of the number of
controller nodes $|Z|$. As expected, for the EM algorithm, when
$|\mathcal{Z}|$ is too small, the FSCs cannot
represent the optimal policy (under-fitting), and when the number of
nodes is too large, FSCs overfits a limited amount of data and perform
poorly. Even if $|\mathcal{Z}|$ is set to the number inferred by Dec-SBPR, EM can still suffer severely from
initialization and local maxima issues, as can be seen from a large
error-bar. By setting a high truncation level ($|Z|=50$), Dec-SBPR 
employs Algorithm~\ref{alg:Dec-SBPR-off-policy} to integrate out the
uncertainty of the policy representation (under the SB prior). As a result, Dec-SBPR 
can infer both the number of nodes that is needed ($\approx5$) and
optimal controller parameters simultaneously. Furthermore, this inference is done with less computation time and with
a higher value and improved robustness (low variance of test value) than EM.
\begin{figure}[!t]
	\begin{center}
		$\begin{array}{cc}
		\hspace{-0.25cm}\includegraphics[width=2.5in]{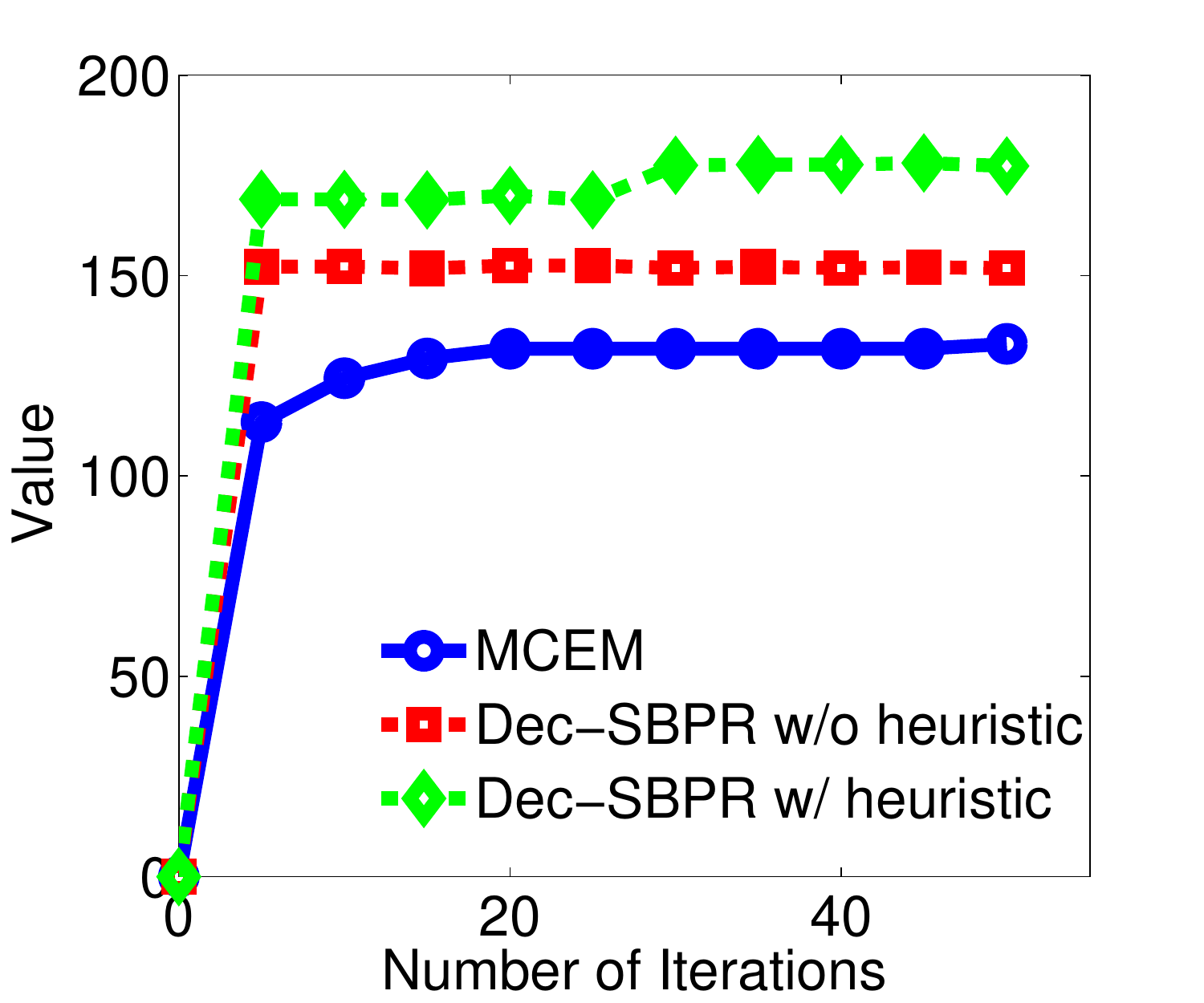}
		\hspace{-0.0cm}\includegraphics[width=2.5in]{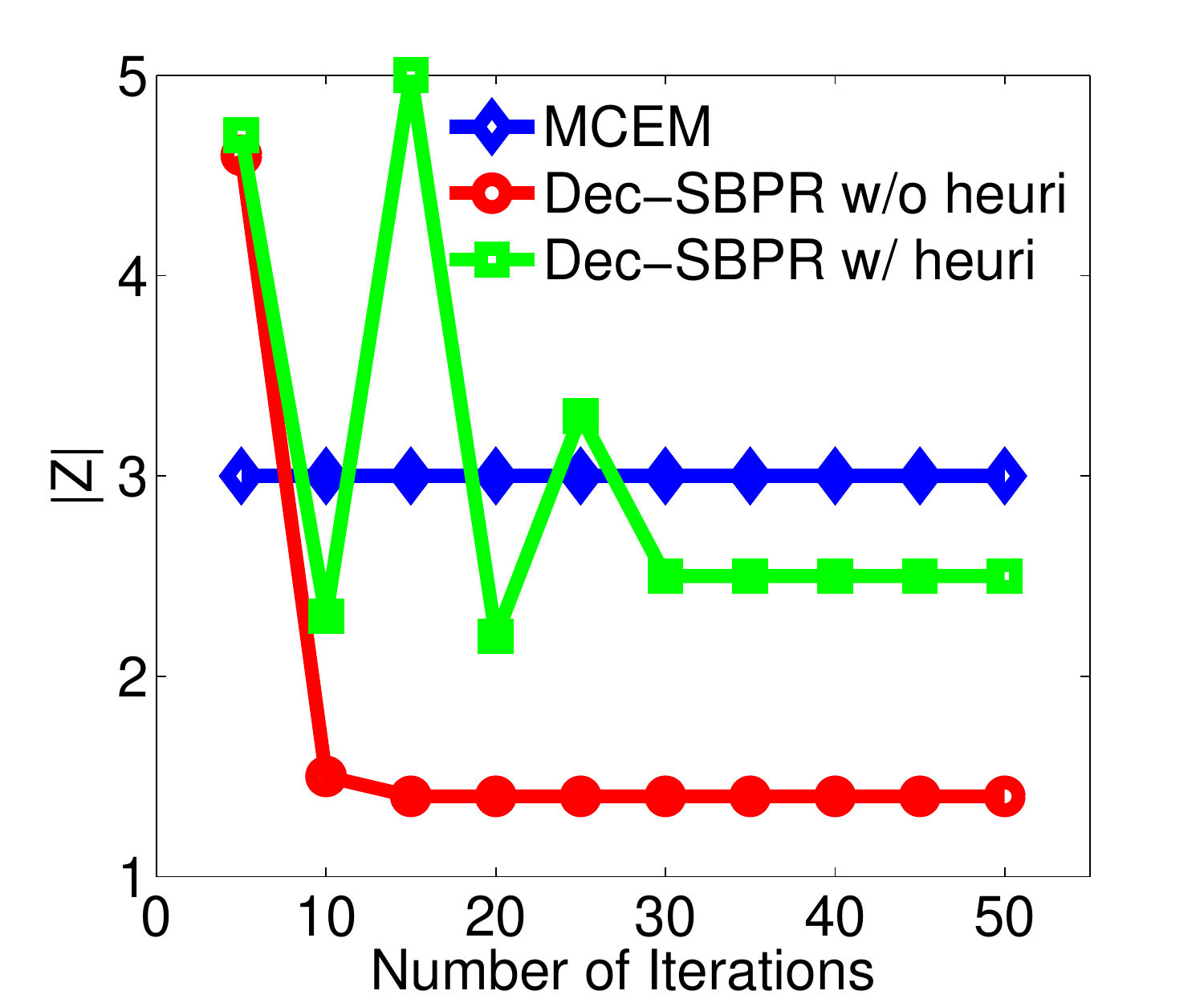}
		\end{array}$
		\vspace{-5pt}
		\caption{\label{fig:traffic}{Performance comparison on 
traffic control problem ($10^{20}$ states and $100$ agents). 
Test reward (left) and inferred controller size (right) of Dec-SBPR, 
as a function of algorithmic iteration.}}
	\end{center}
	\vspace{-0.2in}
\end{figure}

\paragraph{Comparison with other methods} The performance of
Dec-SBPR is also compared to several state-of-art methods, including: Monte
Carlo EM (MCEM)~\cite{WU:IJCAI2013}. Similar to Dec-SBPR, MCEM is a policy-based RL approach. We apply the exploration-exploitation strategy
described in section~\ref{sec:exploration} and follow the same
experimental procedure in \cite{WU:IJCAI2013} to report the
results\footnote{The learning curves of Dec-SBPR are shown in the appendix.}. The rewards after running a fixed number of iterations and a fixed amount of time are summarized (respectively) in Table~\ref{tab:compare} (the first column under policy-learning category). Dec-SBPR is shown to achieve 
better policy values than MCEM on all problems \footnote{The results are provided by
  personal communication with its authors and run on the same benchmarks
  that are available online.}. These results can be explained by the
fact that EM is (more) sensitive to initialization and (more) prone to local
optima. Moreover, by fixing the size of the controllers, the optimal policy
from EM algorithms might be over/under fitted . By using
a Bayesian nonparametric prior, Dec-SBPR learns the policy with
variable-size controllers, allowing more flexibility for representing
the optimal policy. We also show the result of Dec-SBPR running the same amount of clock time as MCEM (Dec-SBPR (fixed time)), which indicates Dec-SBPR can achieve a better trade-off between policy value and learning time than MCEM. 

Finally, Dec-SBPR is compared to  Periodic EM
(PeriEM)~\cite{JJ:NIPS11}
and
FB-HSVI~\cite{dibangoye2014error}, two state-of-art planing methods (with known models)
for generating controllers. Because having a Dec-POMDP model allows more accurate value function calculations than a finite number of trajectories, the
value of PeriEM and FB-HSVI are treated as upper-bounds for
the policy-based methods. Our Dec-SBPR approach can sometimes outperform PeriEM, but produces lower value than FB-HSVI. FB-HSVI is a boundedly-optimal method, showing that Dec-SBPR can produce near optimal solutions in some of these problems and produces solutions that are much closer to the optimal than previous RL methods. 
It is also worth noting that neither PeriEM nor FB-HSVI can scale to large problems (such as the one discussed below), while 
by using a policy-based RL approach, Dec-SBPR can scale well. 
\vspace{-0.15in}
\paragraph{Scaling up to larger domains} To demonstrate scalability to both large problem sizes and large numbers of agents, we test our algorithm on a traffic problem \cite{WU:IJCAI2013}, with $10^{20}$ states. Here, there are $100$ agents controlling the traffic flow at $10\times10$ intersections with one agent located at each intersection. 
Except for MCEM, no previous Dec-POMDPs algorithms are able to solve such large problems.

Since the authors in~\cite{WU:IJCAI2013} use a hand-coded policy (comparing the traffic flow between two directions) as a heuristic for generating training trajectories,  we also use such a heuristic for a fair comparison. In addition, to examine the effectiveness of the exploration-exploitation strategy described in Section \ref{sec:explore-exploit},  we also consider the case where the initial behavior policy is random and then it is optimized as discussed. 
From Figure \ref{fig:traffic}, we can see that, with the help of the heuristic, Dec-SBPR can achieve the best performance. Without using the heuristic (by just using our exploration-exploitation strategy), in a few iterations, Dec-SBPR is able to produce a higher quality policy than MCEM.  Moreover, the inferred number of FSC nodes (averaged over all agents) is smaller than the number preselected by MCEM. This shows that not only can Dec-SBPR scale to large problems, but it can also produce higher-quality solutions than other methods for those large problems.

\section{Conclusions}
The paper presented a scalable Bayesian nonparametric policy
representation and an associated learning framework (Dec-SBPR) for
generating decentralized policies in Dec-POMDPs. An new exploration-exploitation method, which 
extends the popular $\epsilon$-greedy method, was also provided for reinforcement learning in Dec-POMDPs.  Experimental results
show Dec-SBPR produces higher-quality solutions than the
state-of-art policy-based method, and has the additional benefit of
inferring the number of nodes needed to represent the
optimal policy. The resulting method is also scalable to large domains (in terms of both the number of agents and the problem size), allowing high-quality policies for large Dec-POMDPs to be learned efficiently from data.  

\section*{Acknowledgments}

Acknowledgments This research was supported by the US
Office of Naval Research (ONR) under MURI program award \#N000141110688 and NSF award \#1463945.

\appendix
\section*{Appendices}
\addcontentsline{toc}{section}{Appendices}
\renewcommand{\thesubsection}{\Alph{subsection}}
\subsection{Proof of Theorem \ref{theorem:RPR-VB}:(Mean-field) Variational Bayesian (VB) Inference for DEC-SBPR} 
Under the standard variational theory~\cite{beal2003variational, bishop2006pattern}, minimizing the KL divergence between $q(\Theta, z)$ and $p(\Theta, z|\mathcal{D})$ is equivalent to minimizing the lower bound of log marginal likelihood (empirical value function for our case). Using Jensen's inequality, we can obtain the following lower bound of the log marginal value function

\begin{eqnarray}\label{eq:log-marginla-value}
\label{eq:hyper-mu-new1}
&&\hspace{-1.0cm}\ln\widehat{V}(\mathcal{D}^{(K)}) = \ln\frac{1}{K}\sum_{k=1}^K\sum_{t=0}^{T_k}\sum_{\vec{z}_0^k,\cdots,\vec{z}_t^k}\int q_t^k(\vec{z}_{0:t}^k)q(\Theta)q(\eta)\frac{\tilde{r}_t^kp(\Theta)p(\eta)p(\vec{a}^k_{0:t}, \vec{z}_{0:t}^k|\vec{o}^k_{1:t},\Theta)}{q_t^k(\vec{z}_{0:t}^k)q(\Theta)q(\eta)}d\Theta d\eta
\cr\label{eq:q_alpha}
\label{eq:hyper-mu-new2}
&&\hspace{1.0cm} \geq \frac{1}{K}\sum_{k=1}^K\sum_{t=0}^{T_k}\sum_{\vec{z}_0^k,\cdots,\vec{z}_t^k}\int q_t^k(\vec{z}_{0:t}^k)q(\Theta)q(\eta)\ln\frac{\tilde{r}_t^kp(\Theta)p(\eta)p(\rho)p(\vec{a}^k_{0:t}, \vec{z}_{0:t}^k|\vec{o}^k_{1:t},\Theta)}{q_t^k(\vec{z}_{0:t}^k)q(\Theta)q(\eta)}d\Theta d\eta
\cr
\label{eq:hyper-pi-new1}
&&\hspace{1.0cm} = \ln\widehat{V}(\mathcal{D}^{(K)}) - KL\bigg(\big\{q_t^k(\vec{z}_{0:t}^k)q(\Theta)q(\eta)\big\}\big|\big| \frac{\xi_t^k}{\widehat{V}(\mathcal{D}^{(K)})}p(\vec{z}_{0:t}^k,\Theta, \eta|\vec{a}^k_{0:t},\vec{o}^k_{1:t}) \bigg)
\cr\label{eq:hyper-pi-new1}
&&\hspace{1.0cm}\overset{def}{=} \mathrm{LB}\Bigg(\bigg\{\big\{q_t^k(z_{n, 0:t}^k)\big\}_{k, t}, q(\Theta_n), q(\eta_n)\bigg\}_{n=1\cdots N}\Bigg)
\\\label{eq:hyper-W-new}
\nonumber
\end{eqnarray}
where $\xi_t^k=\tilde{r}_t^k\int p(\vec{a}_{0:t}^k|\vec{o}_{1:t}^k, \Theta)p(\Theta)d\Theta$ and $\tilde{r}_t^k=\gamma^tr_t^k/p^\Psi(\vec{a}|\vec{h}_t^k)$. We assume $p(\Theta)=\prod_{n=1}^Np(\Theta)$ and $p(\vec{a}_{0:t},\vec{z}_{0:t}|\vec{o}_{0:t},\Theta) = \prod_{n=1}^Np({a}_{n,0:t},{z}_{n,0:t}|o_{n,0:t},\Theta_n)$ to accommodate decentralized policy representations.  

To derive the VB updating equations, we rewrite the lower bound in equation \eqref{eq:log-marginla-value} as follows
\begin{eqnarray}\label{eq:fictitious-vs-true}
&&\hspace{-0.75cm} \mathrm{LB}\Bigg(\bigg\{\big\{q_t^k(z_{n, 0:t}^k)\big\}_{k, t}, q(\Theta_n), q(\eta_n)\bigg\}_{n=1,\cdots N}\Bigg)
\cr
&&\hspace{-0.75cm} = -\int q(\Theta)\ln\frac{q(\Theta)}{p(\Theta)}d\Theta-\int q(\eta)\ln\frac{q(\eta)}{p(\eta)}d\eta - \frac{1}{K}\sum_{k=1}^K\sum_{t=0}^{T_k}\sum_{\vec{z}_0^k,\cdots,\vec{z}_t^k}q_t^k(\vec{z}_{0:t}^k)\ln\big(q_t^k(\vec{z}_{0:t}^k)\big)
\cr&&\hspace{-0.5cm} \!+\frac{1}{K}\sum_{k=1}^K\sum_{t=0}^{T_k}\sum_{\vec{z}_0^k,\cdots,\vec{z}_t^k}\prod_{n=1}^N\int q_t^k(\vec{z}_{0:t}^k)q(\Theta_n)q(\eta_n)\ln\big(\tilde{r}_t^k\prod_{n=1}^N p_t^k(a_{n,0:t}^k, z_{n,0:t}^k|o_{n,1:t}^k, \Theta_n)\big)d\Theta_nd\eta_n.
\end{eqnarray}
The VB Inference algorithm for DEC-SBPR is based on maximizing $\mathrm{LB}$ w.r.t. the distribution of the joint DEC-SBPR parameters $\big\{\{q_t^k(z_{n, 0:t}^k)\}_{k, t}, q(\Theta_n), q(\eta_n)\big\}_{n=1,\cdots,N}$, which can be achieved by alternating the following steps.\\\\
{\bf Update the distribution of nodes (VB E-step)}: Keeping $\{q(\Theta_n)\}_{n=1,\cdots,N}$ and $\{q(\eta_n)\}_{n=1,\cdots,N}$ fixed, solve $\max_{\{q_t^k(z_{n, 0:t}^k)\}}\mathrm{LB}\Bigg(\bigg\{\big\{q_t^k(z_{n, 0:t}^k)\big\}_{k, t}, q(\Theta_n), q(\eta_n)\bigg\}_{n=1,\cdots,N}\Bigg), \forall n,k,t$ subject to the normalization constraint for $q_t^k(z_{n, 0:t}^k)$. In this step, we construct the Lagrangian
\begin{equation}
F_{q_t^k(z_{n, 0:t}^k)} = \mathrm{LB}\Bigg(\bigg\{\big\{q_t^k(z_{n, 0:t}^k)\big\}_{k, t}, q(\Theta_n), q(\eta_n)\bigg\}_{n=1,\cdots,N}\Bigg) - \lambda\bigg(K-\sum_{k,t,z_{n, 0:t}^k}\prod_{n=1}^Nq_t^k(z_{n, 0:t}^k)\bigg),
\end{equation}
then take derivative w.r.t $q_t^k(z_{n,0:t}^k)$ and set the result to zero

\begin{eqnarray}\label{eq:fic}
&&\hspace{-0.75cm}\frac{\partial F_{q_t^k(z_{n, 0:t}^k)}}{\partial \big(q_t^k(z_{n, 0:t}^k)\big)} = \frac{1}{K}\int p(\Theta_n)\ln\tilde{r}_t^kp(a_{n,0:t}^k, z_{n,0:t}^k|o_{1:t}^k, \Theta_n)d\Theta_n
\cr&&\hspace{-0.5cm} +\frac{1}{K}\sum_{k=1}^K\sum_{t=0}^{T_k}\sum_{\vec{z}_0^k,\cdots,\vec{z}_t^k}\prod_{i\neq n}\int q_t^k(z_{i,0:t}^k)q(\Theta_n)q(\eta_n) \ln\bigg(\tilde{r}_t^k\prod_{n=1}^N p_t^k(a_{n,0:t}^k, z_{n,0:t}^k|o_{n,1:t}^k, \Theta_n)\bigg)d\Theta_nd\eta_n
\cr&&\hspace{-0.5cm}-\frac{1}{K}\sum_{k=1}^K\sum_{t=0}^{T_k}\sum_{\vec{z}_0^k,\cdots,\vec{z}_t^k}\prod_{i\neq n}q_t^k(z_{i,0:t}^k)\ln\bigg(\prod_{n=1}^Nq_t^k(z_{n,0:t}^k)\bigg) -\frac{1}{K}\sum_{k=1}^K\sum_{t=0}^{T_k}\sum_{\vec{z}_0^k,\cdots,\vec{z}_t^k}\prod_{i\neq n}q_t^k(z_{i,0:t}^k)
\cr&&\hspace{-.5cm}+\lambda\sum_{k=1}^K\sum_{t=0}^{T_k}\sum_{\vec{z}_0^k,\cdots,\vec{z}_t^k}\prod_{i\neq n}q_t^k(z_{i,0:t}^k) = 0,
\end{eqnarray}which is solved to give the distribution of nodes $z_{n, 0:t}^k$ for the $n^{th}$ agent
\begin{eqnarray}\label{eq:q_k_t}
&&\hspace{-0.85cm}q_t^k(z_{n, 0:t}^k) = \frac{\tilde{r}_t^k}{C_z}\exp{\bigg\{\int q(\Theta_n)\ln p(a^k_{n, 0:t}, z_{n,0:t}^k|o^k_{n,1:t},\Theta_n)d\Theta_n\bigg\}}
\cr&&\hspace{0.65cm} =\frac{\tilde{r}_t^k}{C_z}\exp{\bigg\{\ln\mu_n^{z_{n,0}^k} + \sum_{\tau=0}^t\langle\ln\pi_{n,z_{n, \tau}^k}^{a_{n,\tau}^k}\rangle_{p(\pi|\hat{\rho})} + \sum_{\tau=1}^t\langle\ln W^{z_{n,\tau-1}^k, z_{n, \tau}^k}_{n,a_{n,\tau-1}^k,o_\tau^k}\rangle_{p(W|\hat{\sigma},\hat{\eta})}\bigg\}}
\cr&&\hspace{0.65cm} =\frac{\tilde{r}_t^k}{C_z}\mu^{z_{n,0}^k}_n\widetilde{\pi}_{n,z_{n,0}^k}^{a_{n,0}^k}\prod_{\tau=1}^t\widetilde{\pi}^{z_{n, \tau}^k}_{n,a_{n,\tau}^k\widetilde{W}^{z_{n, \tau}^k,z_{\tau-1}^k}_{n,a_{\tau-1}^k, o_{\tau}^k}}
\end{eqnarray}
where
\begin{eqnarray}\label{eq:q_theta}
&&\hspace{-0.25cm}\widetilde{\pi}_{a}^{n, i}\!=\!\exp{\bigg\{\langle\ln \pi_a^{n, i}\rangle_{p(\pi|\hat{\rho})}\bigg\}}\!\!=\!\exp{\bigg\{\langle\psi(\hat{\rho}_a^{n,i}) - \psi(\sum_{m=1}^{|\mathcal{A}|}\hat{\rho}_m^{n, i})\rangle\bigg\}}
\cr&&\hspace{-0.75cm}\widetilde{W}^{i, 1}_{n,a,o}\!=\!\exp{\bigg\{\langle\ln W^{i, 1}_{n, a,o} \rangle_{p(W|{\hat{\sigma},\hat{\eta}})}\bigg\}}\!\!=\!\exp{\bigg\{\langle\ln V^{i,1}_{n,a,o} \rangle_{p(V|{\hat{\sigma},\hat{\eta}})}\bigg\}},
\cr&&\hspace{-0.75cm}\widetilde{W}^{i, j}_{n,a,o}\!=\!\exp{\bigg\{\langle\ln W_{n,a,o}^{i, j}\rangle_{p(W|\bm{\hat{\sigma},\hat{\eta}})}\bigg\}}\!\!=\!\exp{\bigg\{\langle\ln V^{i,j}_{n,a,o}\rangle_{p(V|{\hat{\sigma},\hat{\eta}})}  + \sum^{j-1}_{m=1}\langle\ln(1-V^{i,m}_{n,a,o}) \rangle_{p(V|\hat{\sigma},\hat{\eta})}\bigg\}}\!,
\cr&&\hspace{10.75cm}\text{for}\;j\!=\!2,\cdots,\!|\mathcal{Z}_n|\!\!-\!\!1
\cr&&\hspace{-0.75cm}\widetilde{W}^{i, |\mathcal{Z}_n|}_{n, a,o} =\!\exp{\bigg\{\langle\ln W_{n,a,o}^{i, |\mathcal{Z}_n|}\rangle_{p(W|{\hat{\sigma},\hat{\eta}})}\bigg\}} = \exp{\bigg\{ \sum^{|\mathcal{Z}_n|}_{m=1}\langle\ln (1-V_{i,m}^{n,a,o}) \rangle_{p(V|\hat{\sigma},\hat{\eta})}\bigg\}}
\nonumber
\end{eqnarray}
with
\begin{eqnarray}\label{eq:fictitious-vs-true}
&&\hspace{-0.65cm}\langle\ln V^{i,j}_{n,a,o} \rangle_{p(V|\hat{\sigma},\hat{\eta})} = \psi(\sigma^{i,j}_{n,a,o} + \omega^{i,j}_{n,a,o}) - \psi\bigg(\sigma^{i,j}_{n,a,o}+\langle\eta^{i,j}_{n,a,o}\rangle+\sum_{l=j}^{|\mathcal{Z}_n|} \omega^{i,l}_{n,a,o}\bigg),
\cr&&\hspace{-0.65cm}\langle\ln (1-V^{i,j}_{n,a,o}) \rangle_{p(V|\hat{\sigma},\hat{\eta})} = \! \psi\bigg(\langle\Sigma^{i,j}_{n,a,o}\rangle+\sum_{l=j+1}^{|\mathcal{Z}_n|}\omega^{i,l}_{n,a,o}\bigg) - 
											 \psi\bigg(\sigma^{i,j}_{n,a,o}+\langle\Sigma^{i,j}_{n,a,o}\rangle+\sum_{l=j}^{|\mathcal{Z}_n|} \omega^{i,l}_{n,a,o}\bigg),
\cr&&\hspace{-0.65cm}\langle\eta^{i,j}_{n,a,o}\rangle_{p(\eta|\hat{c},\hat{d})} = \hat{c}^{i,j}_{n,a,o}/\hat{d}^{i,j}_{n,a,o}
\nonumber
\end{eqnarray}
where $\psi(\cdot)$ is digamma function, $\omega^{i,j}_{n,a,o}$ is the reward (soft-count) allocated to the transition from state $i$ to $j$, both $\hat{c}^{i,j}_{n,a,o}$ and $\hat{d}^{i,j}_{n,a,o}$ are the posterior parameters of $\eta^{i,j}_{n,a,o}$. 
\\
\\
{\bf Update the sufficient statistics (VB-M Step)}
In VB-M Step, the distribution of nodes $\{q_t^k(\vec{z}_{0\cdots t}^k)\}$, $\forall k,t$ are fixed, the objective is to solve $\max_{q(\Theta_n),q(\eta_n)}\mathrm{LB}\Bigg(\bigg\{\big\{q_t^k(z_{n, 0:t}^k)\big\}_{k, t}, q(\Theta), q(\eta_n)\bigg\}_{n=1,\cdots,N}\Bigg), \forall n$  subject to the normalization constraint that $\int q(\Theta)d\Theta=1$. First we consider finding $q(\Theta_n)$. To that end, we construct the Lagrangian
\begin{equation}
F_{q_{\Theta_n}} = \mathrm{LB}\Bigg(\bigg\{\big\{q_t^k(z_{n, 0:t}^k)\big\}_{k, t}, q(\Theta_n), q(\eta_n)\bigg\}_{n=1,\cdots,N}\Bigg) -\lambda\bigg(1-\int q(\Theta)d\Theta)\bigg).
\end{equation}
Then
\begin{eqnarray}\label{eq:q_theta}
&&\hspace{-0.05cm}\frac{\partial F_{q_{\Theta}}}{\partial \big(q(\Theta)\big)} = \frac{1}{K}\sum_{k,t,z_{n,0:t}^k} \int \prod_{n=1}^Nq(\eta_n)q_t^k(z_{n,0:t}^k)\ln\tilde{r}_t^k\prod_{n=1}^Np(\eta_n)p(a_{n,0:t}^k, z_{n,0:t}^k|o_{1:t}^k, \Theta_n)d\eta
\cr&&\hspace{5cm}- 1 - \ln\frac{q(\Theta_n)}{p(\Theta_n)} + \lambda = 0
\nonumber
\end{eqnarray}
$\;\;\;\;\Rightarrow$
\begin{eqnarray}\label{eq:theta-update}
\label{eq:hyper-mu-new1}
&&\hspace{-0.8cm}q(\Theta) = \prod_{n=1}^N\frac{p(\Theta_n)}{e^{1-\lambda}}\exp\bigg\{\frac{1}{K}\sum_{k,t,z} q_t^k(z_{n, 0:t}^k)\ln p(a^k_{n, 0:t}, z_{n,0:t}^k|o^k_{n,1:t},\Theta_n)\bigg\}
\cr
\label{eq:hyper-mu-new2}
&&\hspace{0cm}=\prod_{n=1}^N\frac{p(\Theta_n)}{e^{1-\lambda}}\exp\bigg\{\frac{1}{K}\sum_{k,t}K\nu_t^k\bigg[\sum_{\tau=1}^t\sum_{i=1}^{|\mathcal{Z}_n|} \phi_{t,\tau}^{n,k}(i)\ln\pi_i^{n, a_{n, \tau}^k} + \sum_{\tau=1}^t\sum_{i,j=1}^{|\mathcal{Z}_n|}\xi_{t, \tau}^{n, k}(i,j)\ln W_{i,j}^{n,a_{\tau-1}^k,o_{\tau}^k} \bigg\}
\cr\label{eq:hyper-pi-new1}
&&\hspace{0cm}=\prod_{n=1}^N\frac{p(\Theta_n)}{e^{1-\lambda}}\prod_{i=1}^{|\mathcal{Z}_n|}\prod_{a=1}^{|\mathcal{A}_n|}\bigg\{\big\{[\pi_{n,a}^i]^{\sum_{k,t,\tau}\nu_t^k\phi_{t,\tau}^k(i)\mathbb{I}(a_{n,\tau}^k,a)}\prod_{j=1}^{|\mathcal{Z}_n|}\prod_{o=1}^{|\mathcal{O}_n|}[W^{i,j}_{n,a,o}]^{\sum_{k,t,\tau}\nu_t^k\xi_{t,\tau-1}^{n,k}(i,j)\mathbb{I}(a_{n,\tau-1}^k,a)\mathbb{I}(o_{n,\tau}^k,o)}\big\}\bigg\}
\cr\label{eq:hyper-W-new}
&&\hspace{-1cm}(\text{\small{from the relation between stick-breaking weights $\bm{p}$ and independent beta random variables in $\bm{V}$}}~\eqref{eq:sbp})
\cr
&&\hspace{0cm}=\prod_{n=1}^N\frac{p(\Theta_n)}{e^{1-\lambda}}\prod_{i=1}^{|\mathcal{Z}_n|}\prod_{a=1}^{|\mathcal{A}_n|}\bigg\{[\pi^a_{n,i}]^{\sum_{k,t,\tau}\nu_t^k\phi_{t,\tau}^{n,k}(i)\mathbb{I}(a,a_\tau^{n,k})}
\cr
&&\hspace{1cm}\times\prod_{j=1}^{|\mathcal{Z}_n|}\prod_{o=1}^{|\mathcal{O}_n|}\big[V^{i,j}_{n,a,o}\prod_{m=1}^{j-1}(1-V^{i,m}_{n,a,o})\big]^{\sum_{k,t,\tau}\nu_t^k\xi_{t,\tau-1}^{n, k}(i,j)\mathbb{I}(a_{n,\tau-1}^k,a)\mathbb{I}(o_{n,\tau}^k,o)}\bigg\}
\cr
&&\hspace{0cm}=\prod_{n=1}^NDir(\hat{{\rho}}_{n,i})\prod_{a=1}^{|\mathcal{A}_n|}\prod_{o=1}^{|\mathcal{O}|_n}GDD(\hat{{\sigma}}_{n,a,o}^{i,:}, \hat{{\eta}}_{n,a,o}^{i,:})\!\!\!\!\!\!\!\!
\end{eqnarray}
where
\begin{eqnarray}\label{eq:theta-update}
\label{eq:hyper-mu-new1}
\label{eq:hyper-pi-new1}
\hspace{-0.0cm}\hat{{\rho}}_{n, i}^{a}&\!\!\!\!=\!\!\!\!&{\rho}^a_{n, i} +\mbox{$\frac{1}{K}\sum_{k=1}^K\sum_{t=0}^{T_k}\sum_{\tau=1}^{t}$}\nu_t^k\phi_{t,\tau-1}^{n, k}(i)\mathbb{I}(a_{\tau-1}^k=a)
\\
\label{eq:hyper-pi-new1}
\hspace{-0.0cm}\hat{{\sigma}}^{i, j}_{n,a,o}&\!\!\!\!=\!\!\!\!& {\sigma}^{i, j}_{n,a,o} + \omega^{i, j}_{n,a,o}
\cr
\hspace{-0.0cm} &\!\!\!\!=\!\!\!\!&{\sigma}^{i, j}_{n,a,o} + \frac{1}{K}\sum_{k=1}^K\sum_{t=0}^{T_k}\sum_{\tau=1}^{t}\nu_t^k\xi_{t,\tau-1}^{n, k}(i,j)\mathbb{I}(a_{\tau-1}^k=a)\mathbb{I}(o_\tau^k=o)
\cr\label{eq:hyper-W-new}
\hspace{-0.0cm}\hat{{\eta}}^{i, j}_{n,a,o}&\!\!\!\!=\!\!\!\!&{\eta}^{i, j}_{n,a,o} + \sum_{l=j+1}^{|\mathcal{Z}_n|}\omega^{i, l}_{n,a,o}
\cr
\hspace{-0.0cm}&\!\!\!\!=\!\!\!\!& {\eta}^{i, j}_{n,a,o} + \frac{1}{K}\sum_{l=j+1}^{|\mathcal{Z}_n|}\sum_{k=1}^K\!\sum_{t=0}^{T_k}\hat{\nu}^k_t\sum_{\tau=1}^{t}
\xi_{t,\tau-1}^{n, k}(i,j)\mathbb{I}(a_{\tau-1}^k=a)\mathbb{I}(o_{\tau}^k,o),\nonumber
\end{eqnarray}
and $\nu^{\,k}_t$ is the marginalized re-weighted reward computed by equation~\ref{eq:reweightedreward}.\\

To find $q(\eta)$, we construct the Lagrange,
\begin{equation}
F_{q(\Sigma)} = \mathrm{LB}\big(\{q_t^k\}, g(\Theta), q(\eta)\big) -\lambda\bigg(1-\int q(\eta)d\eta)\bigg)
\end{equation}

\begin{equation}\label{eq:q_sigma-vb}
\begin{aligned}
\frac{\partial F_{\eta}}{\partial \big(q(\eta)\big)} = \frac{1}{K} \int q(\Theta)\ln p(\eta)p(\Theta) d\Theta - \frac{1}{K}\ln q(\eta) + \lambda = 0
\end{aligned}
\end{equation}
$\;\;\;\;\Rightarrow$
\begin{eqnarray}
&&\hspace{-0.75cm}q(\eta) \propto \exp\bigg\{\int q(\Theta)\ln p(\eta)p(\Theta|\eta) d\Theta \bigg\}
\cr
&&\hspace{.25cm} \propto \exp\bigg\{\int q(\Theta)\ln p(\Theta|\eta) d\Theta + \ln p(\eta) \bigg\}
\cr
&&\hspace{.25cm} \propto p(\eta) \exp\big\{\langle\ln p(\Theta)\rangle\big\}
\\
&&\hspace{.25cm}=\prod_{n=1}^{N}\prod_{a=1}^{|\mathcal{A}_n|}\prod_{j=1}^{|\mathcal{O}_n|}\prod_{i=1}^{|\mathcal{Z}_n|}\prod_{j=1}^{|\mathcal{Z}_n|}\mathrm{Ga}(\eta_{i,j}^{n,a,o};c,d)\frac{\Gamma(\sigma_{i,j}^{n,a,o}+\eta_{i,j}^{n,a,o})}{\Gamma(\sigma_{i,j}^{n,a,o})\Gamma(\eta_{i,j}^{n,a,o})}{V_{i,j}^{n,a,o}}^{^{\sigma_{i,j}^{n,a,o}-1}}(1-V_j^{n,a,o})^{\eta_{i,j}^{n,a,o}-1}
\cr
&&\hspace{0.25cm}\approx\prod_{n=1}^{N}\prod_{a=1}^{|\mathcal{A}_n|}\prod_{j=1}^{|\mathcal{O}_n|}\prod_{i=1}^{|\mathcal{Z}_n|}\prod_{j=1}^{|\mathcal{Z}_n|}\frac{\Gamma(\sigma_{i,j}^{n,a,o}+\eta_{i,j}^{n,a,o})}{\Gamma(\eta_{i,j}^{n,a,o})}\big\{\eta_{i,j}^{n,a,o}\big\}^{c-1}\exp\big\{-\eta_{i,j}^{n,a,o}(d-\ln(1-V_j^{n,a,o}))\big\}
\cr\label{eq:hyper-pi-new1}\nonumber
\end{eqnarray}
One can set $\sigma_{i,j}^{n,a,o} = 1$, in this case, the VB approximation of $q(\sigma)$ is a product of independent gamma distributions. However, 
when $\sigma_{i,j}^{n,a,o}\neq 1$, \eqref{eq:q_sigma-vb} is no longer a gamma distribution (the prior and likelihood are not conjugate). To solve this issue, one might consider the VB inference method for non-conjugate priors \cite{VBnonconjugate}, by which we consider a point estimate of $\eta$, such at $q(\eta)$ is maximized. One way to obtain the maximum estimate of $\eta$ is to solve $\frac{\partial q(\eta)}{\partial \eta} = 0$, however this operation involves taking derivative w.r.t gamma functions, which does not have a simple form solution. To circumvent this difficult, we use grid search. To make the search more efficient, we use the bounds of $\frac{\Gamma(a+x)}{\Gamma(x)}$ to give an initial estimate of the searching range. The bounds are from Wendel's double Inequality~\cite{qi2010bounds}.
\begin{equation}
x(x+a)^{a-1} \leq \Gamma(a+x)/\Gamma(x) \leq x^{a}
\end{equation}
where $x, a>0$. 
	
\subsection{Some Basics of Stick-breaking Priors}
Here we provide the definition of stick-breaking prior (SBP), its connection to generalized Dirichlet distribution and the corresponding posterior inference, as well as the main statistics characteristics which are useful for developing the inference methods in our paper. For more detailed mathematical treatment, readers are referred to \cite{ishwaran2001gibbs, wong1998generalized}
\begin{definition}{}
The stick-breaking priors \cite{ishwaran2001gibbs} are almost surely discrete random probability measures $\mathcal{P}$ over the the measurable space $(\Omega, \mathcal{B})$ which are partitioned into $d$ disjoint regions with $\Omega=\cup\mathcal{B}_k$ for $1,\cdots,d$. It is expressed as 
\begin{equation}
\mathcal{P}^d_{a,b} = \sum_{k=1}^dp_i\delta_{\Theta_i}
\end{equation}
and
\begin{equation}\label{eq:sbp}
p_1 = V_1\;\;\;\;\text{and}\;\;\;\;p_i = (1-V_1)(1-V_2)\cdots(1-V_{i-1}) V_i, i>2
\end{equation}
are the weights with $V_i$ are independent $Beta(a_k, b_k)$ random variables for $a_i, b_i>0$.
\end{definition}
SBP allows $\mathrm{Beta}$-distributed RVs $V_i,\forall i$ and the atoms $\Theta_i, \forall i$ associated with the resulting weights to be drawn simultaneously. 

\subsubsection{Stick-breaking prior and generalized Dirichlet distribution}
We denote $\bm{p}\sim SB(\bm{v},\bm{w})$ as constructing $\bm{p}$ as an infinite process ($d\rightarrow\infty$) as \eqref{eq:sbp}, and $\bm{p}\sim GDD(\bm{v},\bm{w})$ when $\bm{p}$ is finite. Here, GDD stands for generalized Dirichlet distribution. To see the connection between SBP and GDD, set the truncation level (number of occupied states) to $d$ with $p_{d+1} = 1 - \sum_{i=1}^d p_i$, then we can write down the density function of $\bm{V}=(V_1,\cdots,V_d)$ as 
\begin{equation}
f(\bm{V}) = \prod_{i=1}^df(V_i)=\prod_{i=1}^d\frac{\Gamma(v_i+w_i)}{\Gamma(v_i)\Gamma(w_i)}V_i^{v_i-1}(1-V_i)^{w_i-1}.
\end{equation}
By changing variables from $\bm{V}$ to $\bm{p}$ and using the relation between $\bm{V}$ and $\bm{p}$ as described by~\eqref{eq:sbp}, the density of $\bm{p}$ can be obtained as follows,
\begin{equation}\label{eq:gdd}
\begin{aligned}
f(\bm{p}) = |\frac{\partial \bm{V}}{\partial \bm{p}}| f(\bm{V}) = &\prod_{i=1}^d\bigg(\frac{\Gamma(v_i+w_i)}{\Gamma(v_i)\Gamma(w_i)}p_i^{v_i-1}\big(1-\sum_{j=1}^ip_j\big)^{w_i-(v_{i+1}+w_{i+1})}\bigg)p_{d+1}^{w_d-1}
\end{aligned}
\end{equation}
which has a mean and variance for an element $p_i$,
\begin{equation}
\begin{aligned}
\mathbb{E}[p_i] =& \frac{v_i\prod_{l=1}^{i-1}w_l}{\prod_{l=1}^i(v_l+w_l)},\\
 \mathbb{V}[p_i] =& \frac{v_i(v_i+1)\prod_{l=1}^{i-1}w_l(w_l-1)}{\prod_{l=1}^i(v_l+w_l)(v_l+w_l+1)}.
\end{aligned}
\end{equation}
When $w_i=\sum_{j=i+1}^Kv_i$ for $i<d$, and keeping $w_d=w_d$, the GDD is equivalent to the standard Dirichlet distribution. 

As a concrete example, consider the case $d=3$, we have  
\begin{equation*}
\begin{aligned}[c]
      p_1 &= V_1\\
      p_2 &= (1-V_1)V_2\\
      p_3 &= (1-V_1)(1-V_2)V_3\\
      p_4 &= (1-V_1)(1-V_2)(1-V_3)V_4\\
      &= (1-V_1)(1-V_2)(1-V_3)\
\end{aligned}
\qquad\Longleftrightarrow\qquad
\begin{aligned}[c]
V_1 &= p_1\\
V_2 &= \frac{p_2}{1-p_1}\\
V_3 &= \frac{p_3}{1-p_1-p_2}\\
V_4 &= \frac{p_4}{1-p_1-p_2-p_3} = 1.\\
\end{aligned}
\end{equation*}
Plugging these relations into (\ref{eq:gdd}), we can obtain
\begin{eqnarray}
\label{eq:hyper-mu-new1}
&&\hspace{-0.75cm}f(\bm{p}) =  |\frac{\partial \bm{V}}{\partial \bm{p}}| f(\bm{V})
\cr
&&\hspace{-0.0cm} = \prod_{i=1}^4(1-\sum_{j=1}^{i-1}p_j)\frac{\Gamma(v_i+w_i)}{\Gamma(v_i)\Gamma(w_i)}\times p_1^{v_1-1}(1-p_1)^{w_1-1}\cdot(\frac{p_2}{1-p_1})^{v_2-1}(1-\frac{p_2}{1-p_1})^{w_2-1}
\\
\label{eq:q_alpha}
\label{eq:hyper-mu-new2}
&&\hspace{-0.0cm} =  (\frac{p_3}{1-p_1-p_2})^{v_3-1}(1-\frac{p_2}{1-p_1-p_2})^{w_3-1}\cdot(\frac{p_4}{1-p_1-p_2-p_3})^{v_4-1}(1-\frac{p_4}{1-p_1-p_2-p_3})^{w_4-1}
\cr
\label{eq:hyper-pi-new1}
&&\hspace{-0.0cm}=\prod_{i=1}^4 \frac{\Gamma(v_i+w_i)}{\Gamma(v_i)\Gamma(w_i)}p_i^{v_i-1}(1-\sum_{j=1}^ip_j)^{w_i-(v_{i+1}+w_{i+1})}.
\label{eq:hyper-W-new}
\nonumber
\end{eqnarray}

\subsubsection{Bayesian Inference for GDD}
Given a set of discrete observations $\{X_n\}\overset{i.i.d.}{\sim} Discrete(\bm{p})$, and the prior $\bm{p}\sim SB(v, w)$, the posterior of $\bm{p}$ can be written down as 
\begin{eqnarray}
\label{eq:hyper-mu-new1}
&&\hspace{-0.75cm}p(\bm{p}|\{X_n\}) \propto \prod_{n=1}^N \prod_{i=1}^d p_i^{\mathbb{I}(X_n, i)}\prod_{i=1}^d\bigg(\frac{\Gamma(v_i+w_i)}{\Gamma(v_i)\Gamma(w_i)}p_i^{v_i-1}\big(1-\sum_{j=1}^ip_j\big)^{w_i-(v_{i+1}+w_{i+1})}\bigg)p_{d+1}^{w_d-1}
\cr
&&\hspace{-0.0cm} \propto \prod_{i=1}^d {\big(V_i\prod_{j=1}^{i-1}(1-V_j)\big)}^{\sum_{n=1}^N\mathbb{I}(X_n, i)}V_i^{v_i-1}(1-V_i)^{w_i-1}
\cr\label{eq:q_alpha}
\label{eq:hyper-mu-new2}
&&\hspace{-0.0cm} \propto \prod_{i=1}^dV_i^{v_i+{\sum_{n=1}^N\mathbb{I}(X_n, i)}-1}(1-V_i)^{w_i+\sum_{j>i}\sum_{n=1}^N\mathbb{I}(X_n, i)-1}
\cr
&&\hspace{-0.0cm} = GDD(\bm{v}', \bm{w}')
\\\label{eq:hyper-pi-new1}
\cr\label{eq:hyper-W-new}
\nonumber
\end{eqnarray}
\vskip-0.5in
where the posterior hyper-parameters are updated as $v'_i = v_i + \sum_{n=1}^N\mathbb{I}(X_n=i)$ and $w'_i = w_i + \sum_{j>i}\sum_{n=1}^N\mathbb{I}(X_n=i)$, where $\mathbb{I}(\cdot)$ is an indicator function with value equal to one when the argument is true and zero otherwise. 

\subsection{The Computation of Forward and Backward Variables $(\alpha, \beta)$}
The forward and backward variables $(\alpha,\beta)$ $\alpha_{\tau}^{n, k}(i)\!\!=\!\!p\big(z_{n, \tau}^{k}\!\!=\!\!i | a_{n, 0:\tau}^{\,k}, o_{n, 1:\tau}^{\,k}, {\Theta}_n\big)$ and $ \beta_{t,\tau}^{n, k}(i)\!\!=\!\!\frac{p(a_{n, \tau+1:t}^{\,k} | z_{n, \tau}^{\,k}=i, o_{n, \tau+1:t}^{\,k}, {\Theta}_n)}{\prod_{\tau'=\tau}^t p(a_{\tau}^{\,k}| h_{n, \tau'}^{\,k}, {\Theta}_n)}$, $\forall\,n,k,t,\tau$ defined in section~\ref{sec:vbsb_inference} are similar to the forward-backward messages in the Baum-Welch algorithm for hidden Markov models \cite{rabiner1989tutorial}. These variables are computed recursively by each agent using \eqref{eq:alpha-forward}-\eqref{eq:pah}.

\begin{equation}\label{eq:alpha-forward}
\alpha_{\tau}^{n,k}(i) = \left\{ 
\begin{array}{l l}
\frac{\mu(z_{n, 0}^{\,k} = i)\pi(z_{n, 0}^{\,k}=i,a_{n,0}^{\,k})}{p(a_{ n, 0}^{\,k}|h_{n, 0}^{\,k}, \Theta_n)} & \quad \text{if $\tau=0$}\\
\frac{\sum_{j=1}^{|\mathcal{Z}_n|}\alpha(z_{i,\tau-1}^{\,k}(j)W(z_{i, \tau-1}^{\,k} = j, a_{n,\tau-1}^{\,k}, o_{n,\tau}^{\,k}, z_{n,\tau}^{\,k} = i)\pi(z_{n,\tau}^{\,k} =i ,a_{n,\tau}^{\,k})}{p(a_{n, \tau}^{\,k}|h_{n, \tau}^{\,k}, \Theta_n)} & \quad \text{if $\tau> 0$}
\end{array} \right.
\end{equation}

\begin{equation}\label{eq:beta-backward}
\beta_{t, \tau}^{n, k}(j) = \left\{ 
\begin{array}{l l}
\frac{1}{p(a_{n, 0}^{\,k}|h_{n,0}^{\,k}, \Theta_n)} & \quad \text{if $\tau=0$}\\
\frac{\sum_{n=1}^{|\mathcal{Z}_n|}W(z_{n, \tau}^{\,k} = i, a_{n,\tau-1}^{\,k}, o_{n,\tau}^{\,k}, z_{n,\tau+1}^{\,k}=j)\pi(z_{n,\tau+1}^{\,k}=j ,a_{n,\tau+1}^{\,k})\beta_{n, t, \tau+1}^{\,k}(j)}{p(a_{n, \tau}^{\,k}|h_{n, \tau}^{\,k}, \Theta_n)} & \quad \text{if $\tau> 0$}
\end{array} \right.
\end{equation}

\begin{equation}\label{eq:pah}
p(a_{n, \tau}^{\,k}|h_{n, \tau}, \Theta_n) = \left\{ 
\begin{array}{l l}
\sum_{j=1}^{|\mathcal{Z}_n|}\mu(z_{n,0}^{\,k}=i)\pi(z_{n,0}^{\,k}=i,a_{n,0}^{\,k}) \;\;\;\;\;\;\;\;\;\;\;\;\;\;\;\;\;\;\;\;\;\;\;\;\;\;\;\;\;\;\;\;\;\;\;\;\;\;\;\;\;\;\; \quad \text{if $\tau=0$}\\
\sum_{i,j=1}^{|\mathcal{Z}_n|}\alpha(z_{n,\tau-1}^{\,k}=i)W(z_{n, \tau-1}^{\,k}=i, a_{n,\tau-1}^{\,k}, o_{n,\tau}^{\,k}, z_{n,\tau}^{\,k}=j)\pi(z_{n,\tau}^{\,k}=j, a_{n,\tau}^{\,k}) \\\;\;\;\;\;\;\;\;\;\;\;\;\;\;\;\;\;\;\;\;\;\;\;\;\;\;\;\;\;\;\;\;\;\;\;\;\;\;\;\;\;\;\;\;\;\;\;\;\;\;\;\;\;\;\;\;\;\;\;\;\;\;\;\;\;\;\;\;\;\;\;\;\;\;\;\;\;\;\;\;\;\;\;\;\;\;\;\;\;\;\;\; \quad \text{if $\tau> 0$}
\end{array} \right.
\end{equation}

\subsection{Additional Experimental results}
\subsubsection{learning variable-size FSC vs learning fixed-size FSC}
Additional experiments are added to study the impact of number of training samples.
\begin{figure}[H]
\centering
\hspace{-0.9cm}
\subfigure{
\includegraphics[scale=0.35]{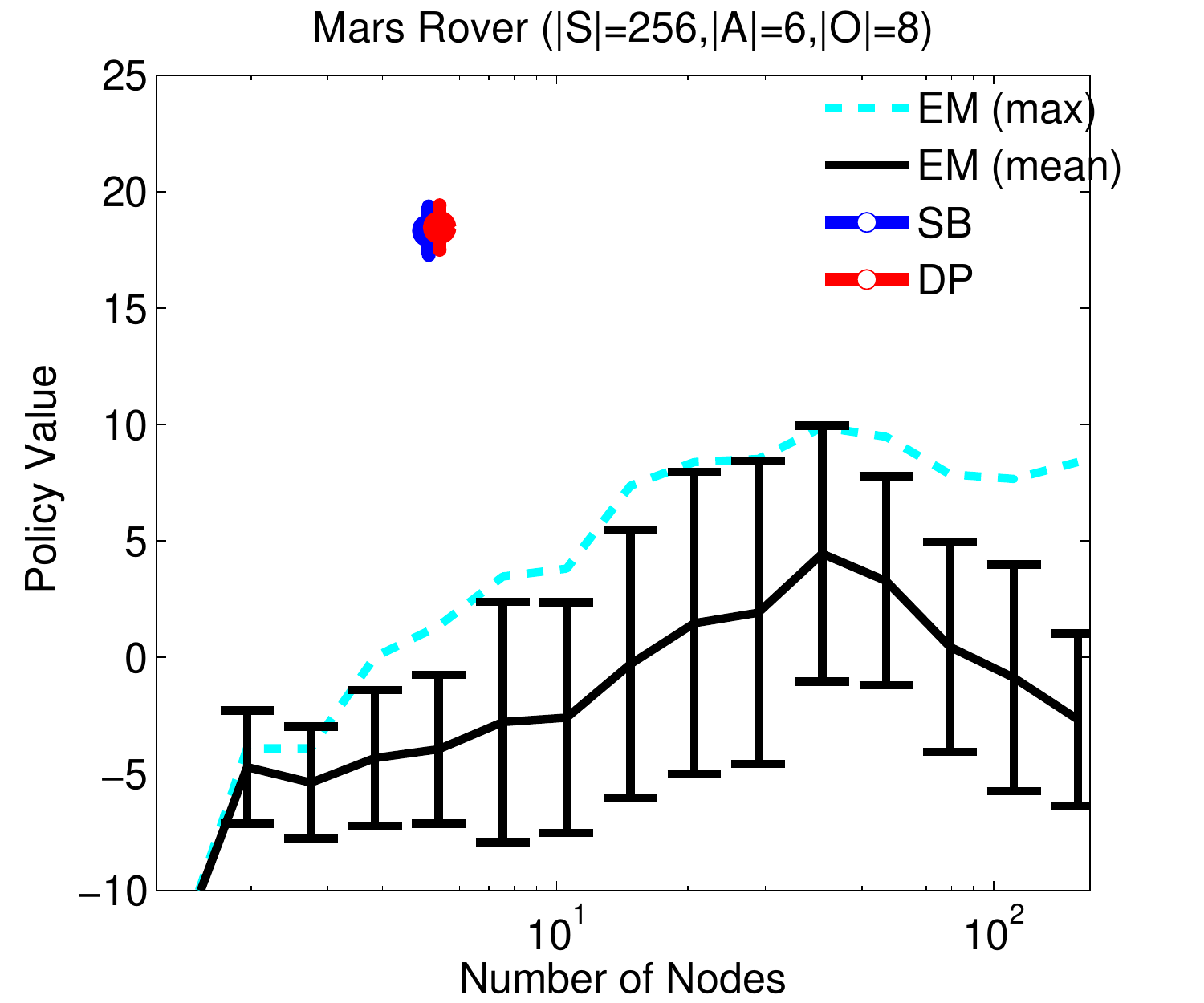}
\label{fig:RoverK30}
}
\hspace{-0.75cm}
\subfigure{
\includegraphics[scale=0.35]{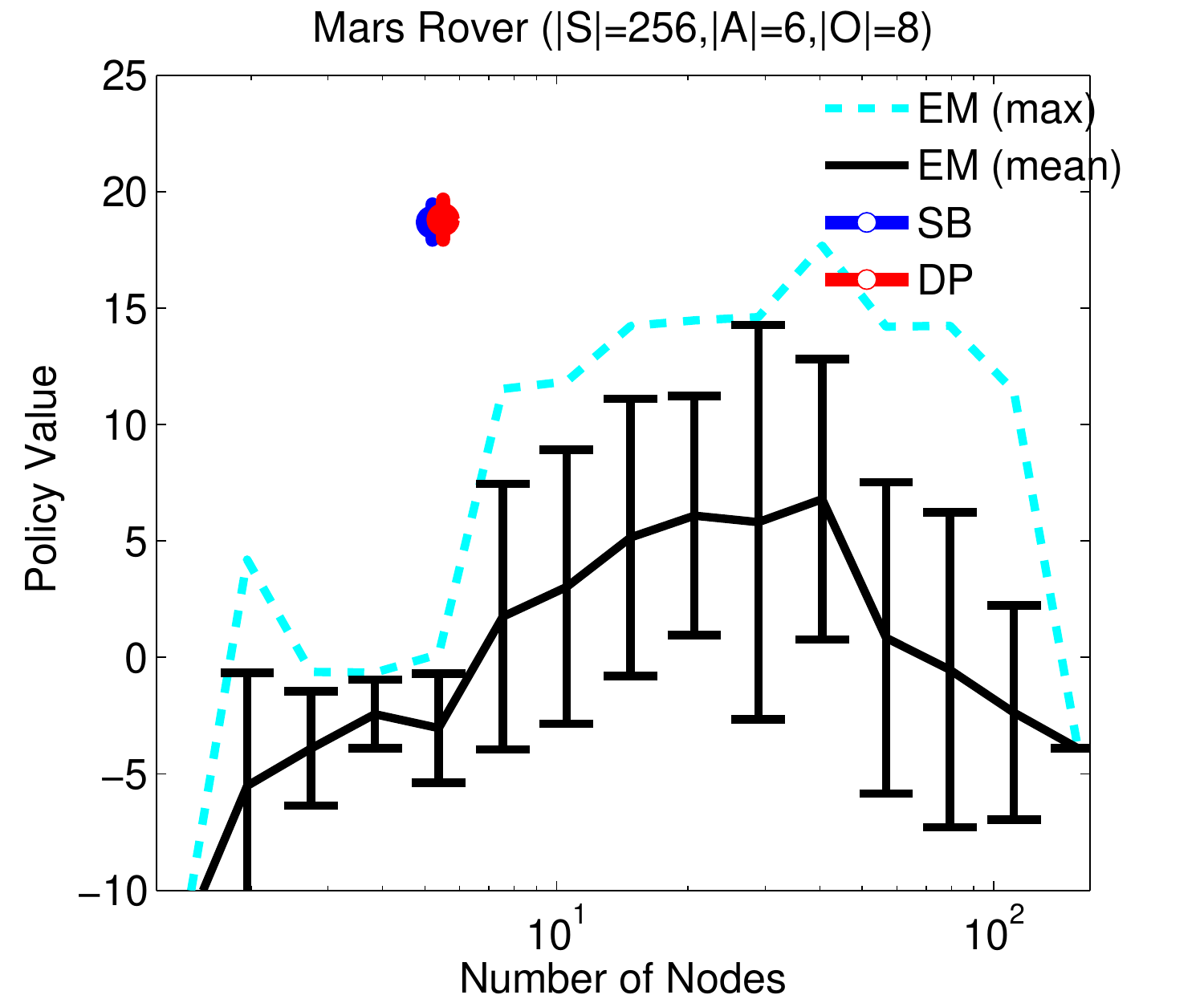}

\label{fig:RoverK100}
}
\hspace{-0.75cm}
\subfigure{
\includegraphics[scale=0.35]{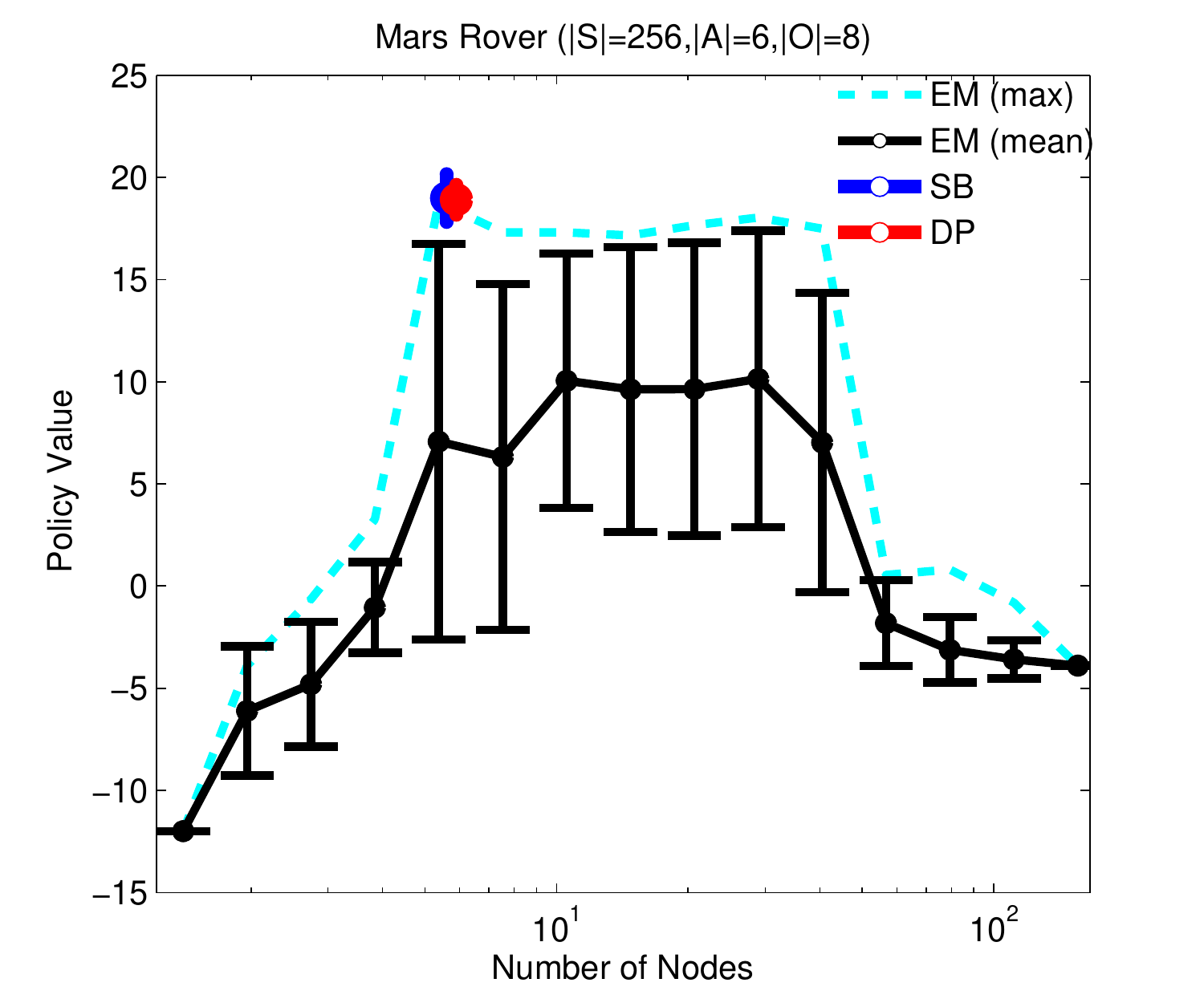}

\label{fig:RoverK300}
}
\vskip-0.1in
\hspace{-0.75cm}
\subfigure{
\includegraphics[scale=0.35]{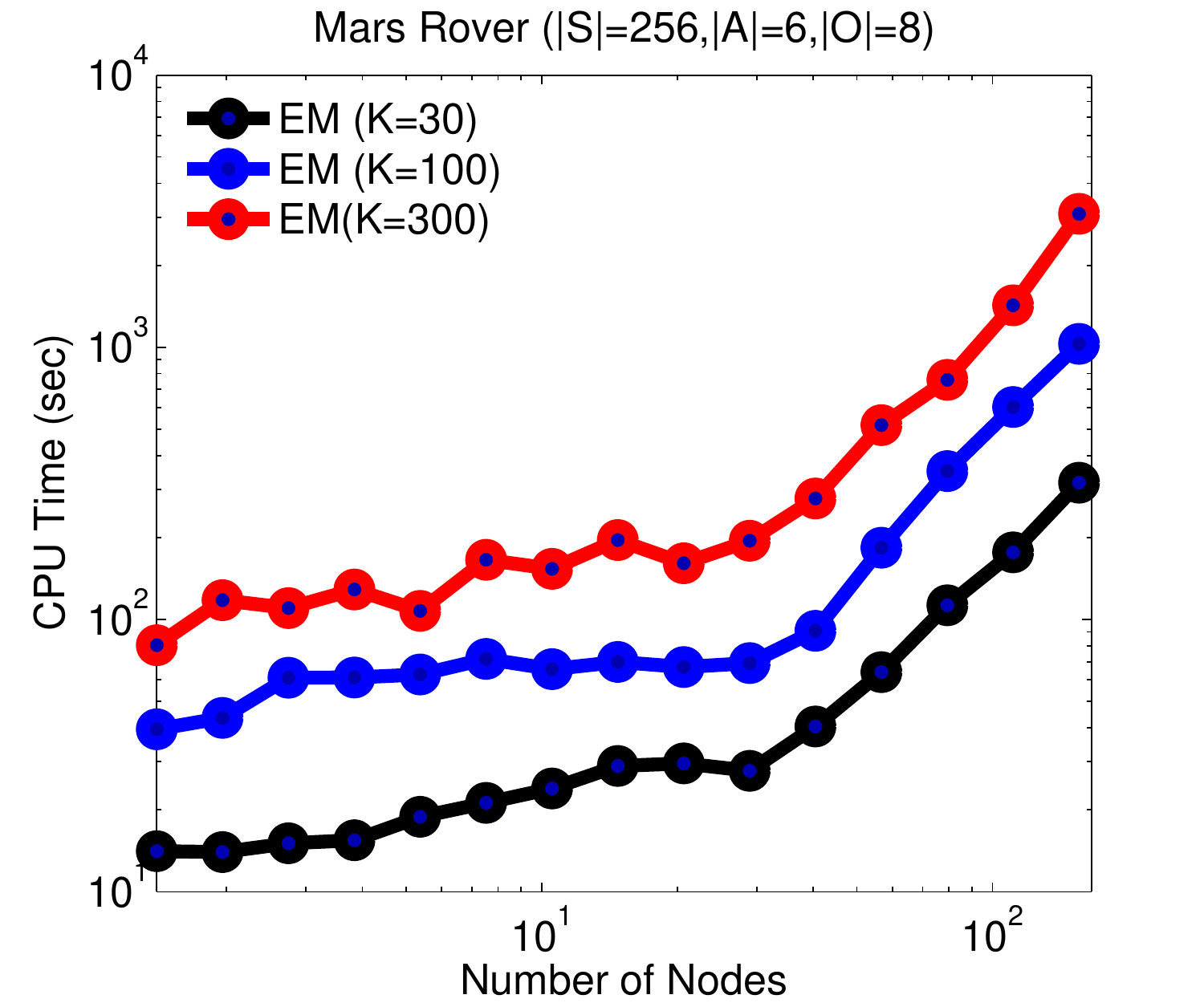}
\label{fig:EMtime}
}
\hspace{-0.75cm}
\subfigure{
\includegraphics[scale=0.35]{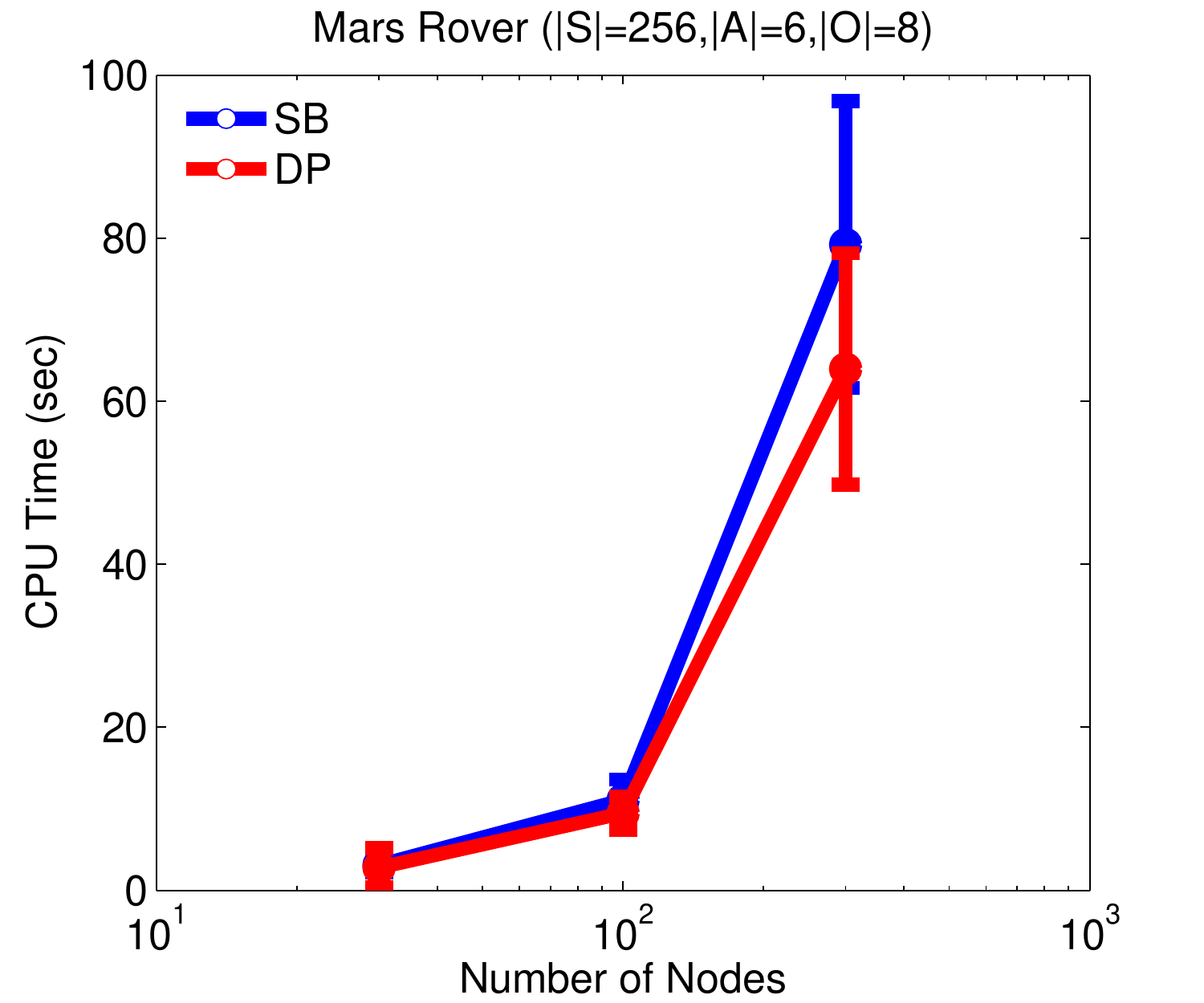}
\label{fig:SBtime}
}
\vskip-0.2in
\caption{ Comparison between variable-size controller learned by Dec-SBPR and fixed-size controller learned by EM algorithm. Top: testing value of policies learned using different number of training episodes($K=30$(left),$K=100$(middle),$K=300$(right)); Bottom: averaged computation time of EM (left) and SB (right).}
\label{fig:additional plots}
\end{figure}
\subsubsection{Sequential Batch Learning with Exploration Exploitation Trade-offs}
\begin{figure}[H]
\centering
\hspace{-0.75cm}
\subfigure{
\includegraphics[scale=0.35]{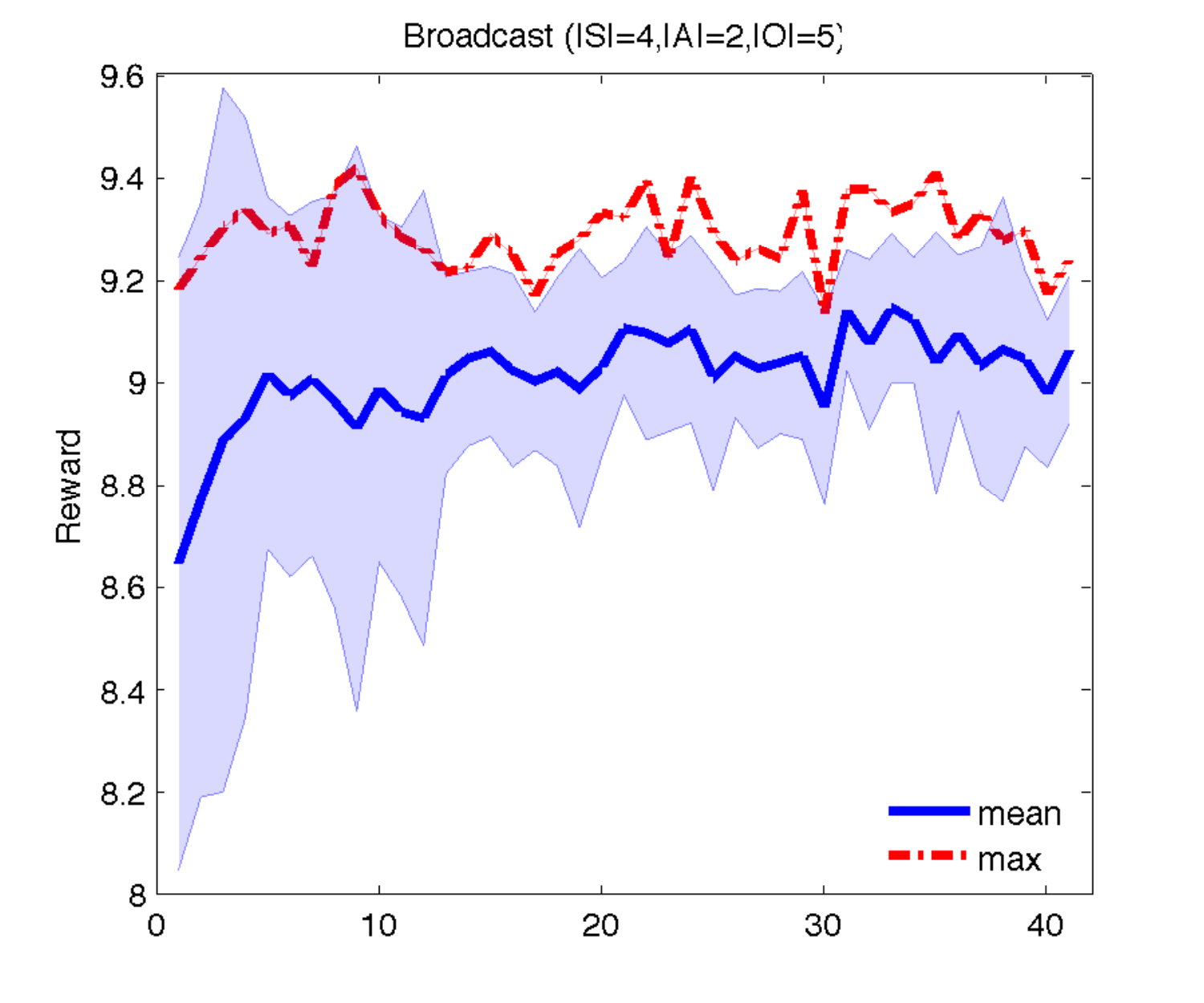}
\label{fig:EMtime}
}
\hspace{-0.75cm}
\subfigure{
\includegraphics[scale=0.35]{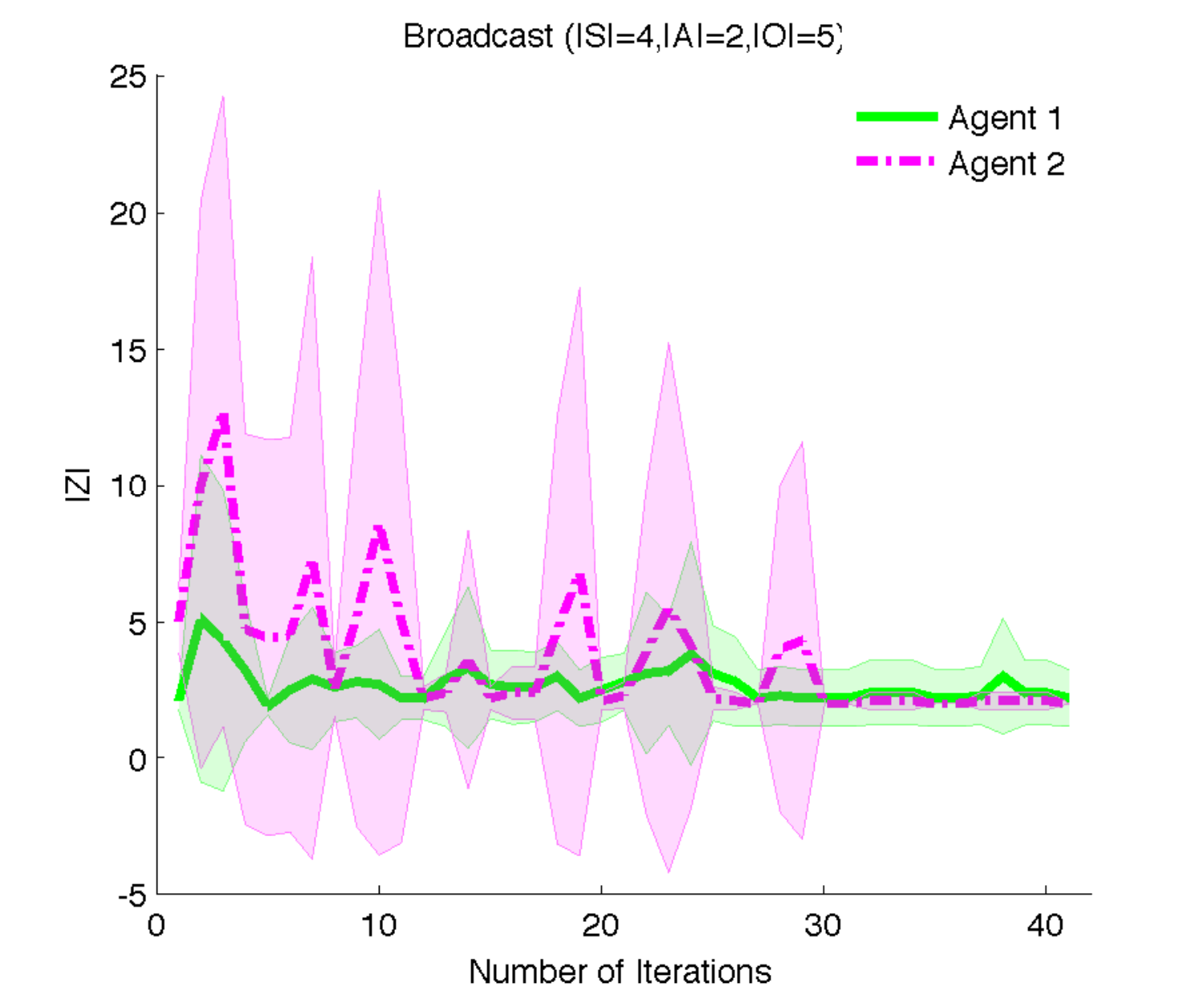}
\label{fig:SBtime}
}
\hspace{-0.75cm}
\subfigure{
\includegraphics[scale=0.35]{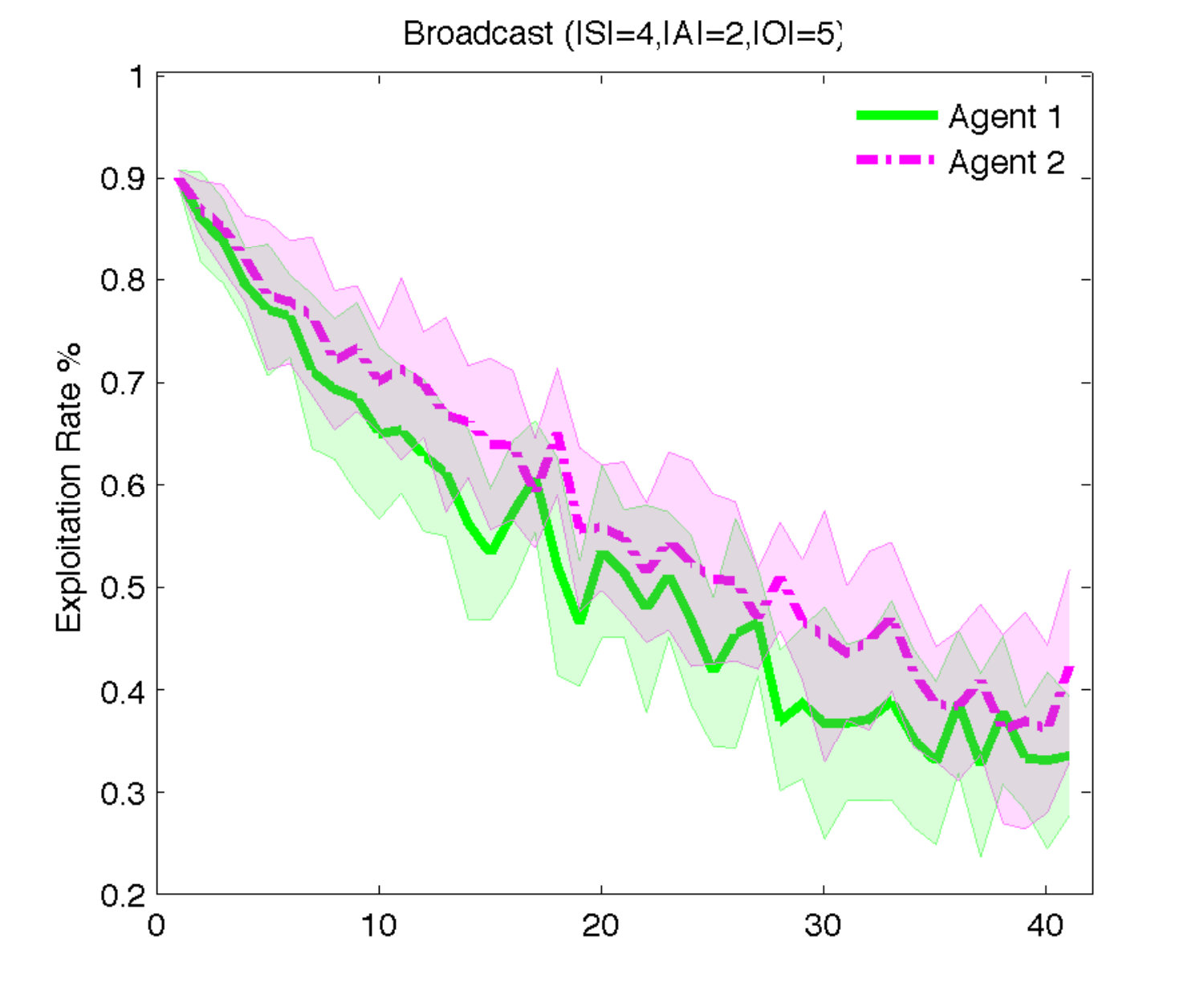}
\label{fig:SBtime}
}
\hspace{-0.75cm}
\subfigure{
\includegraphics[scale=0.35]{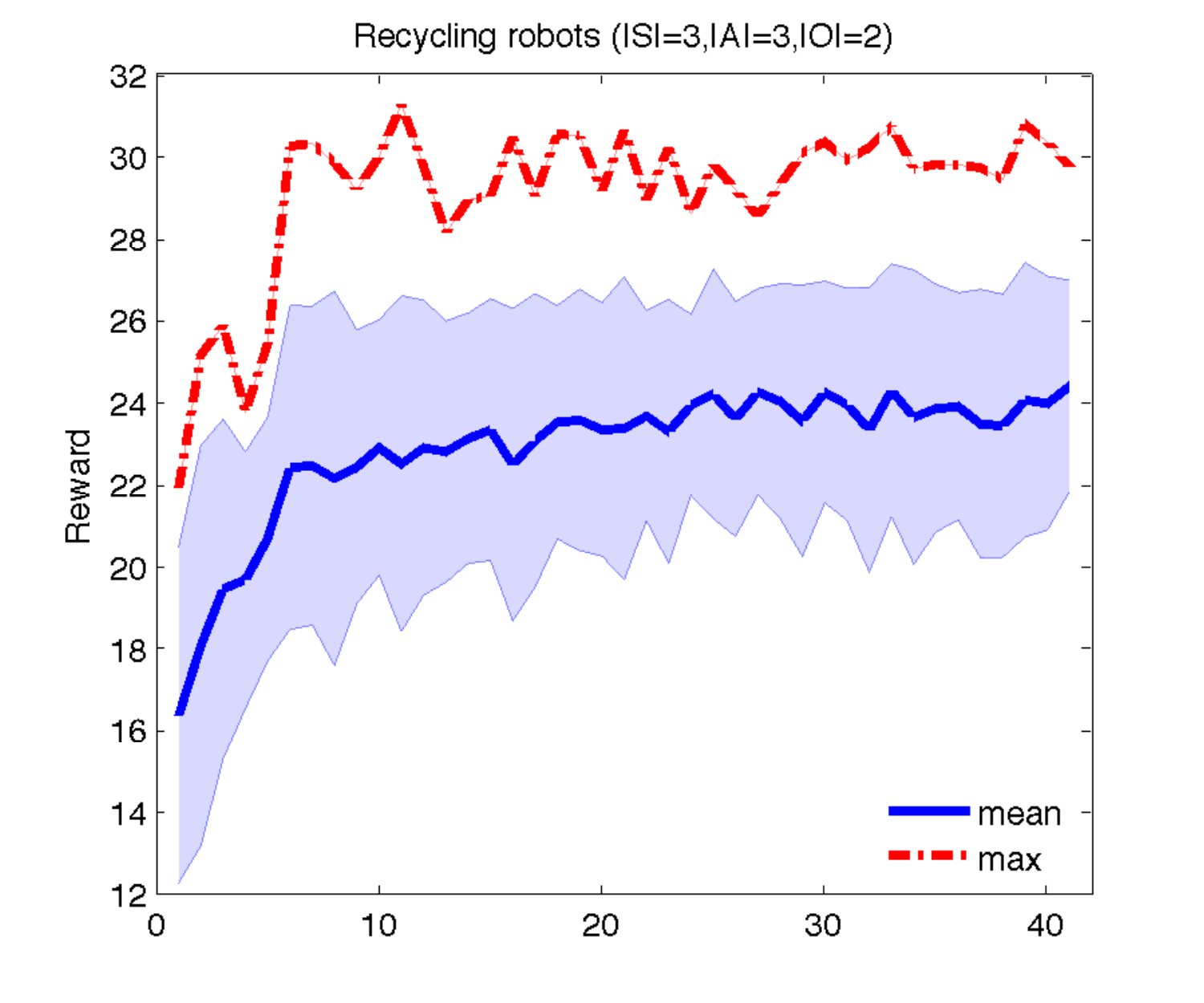}
\label{fig:EMtime}
}
\hspace{-0.75cm}
\subfigure{
\includegraphics[scale=0.35]{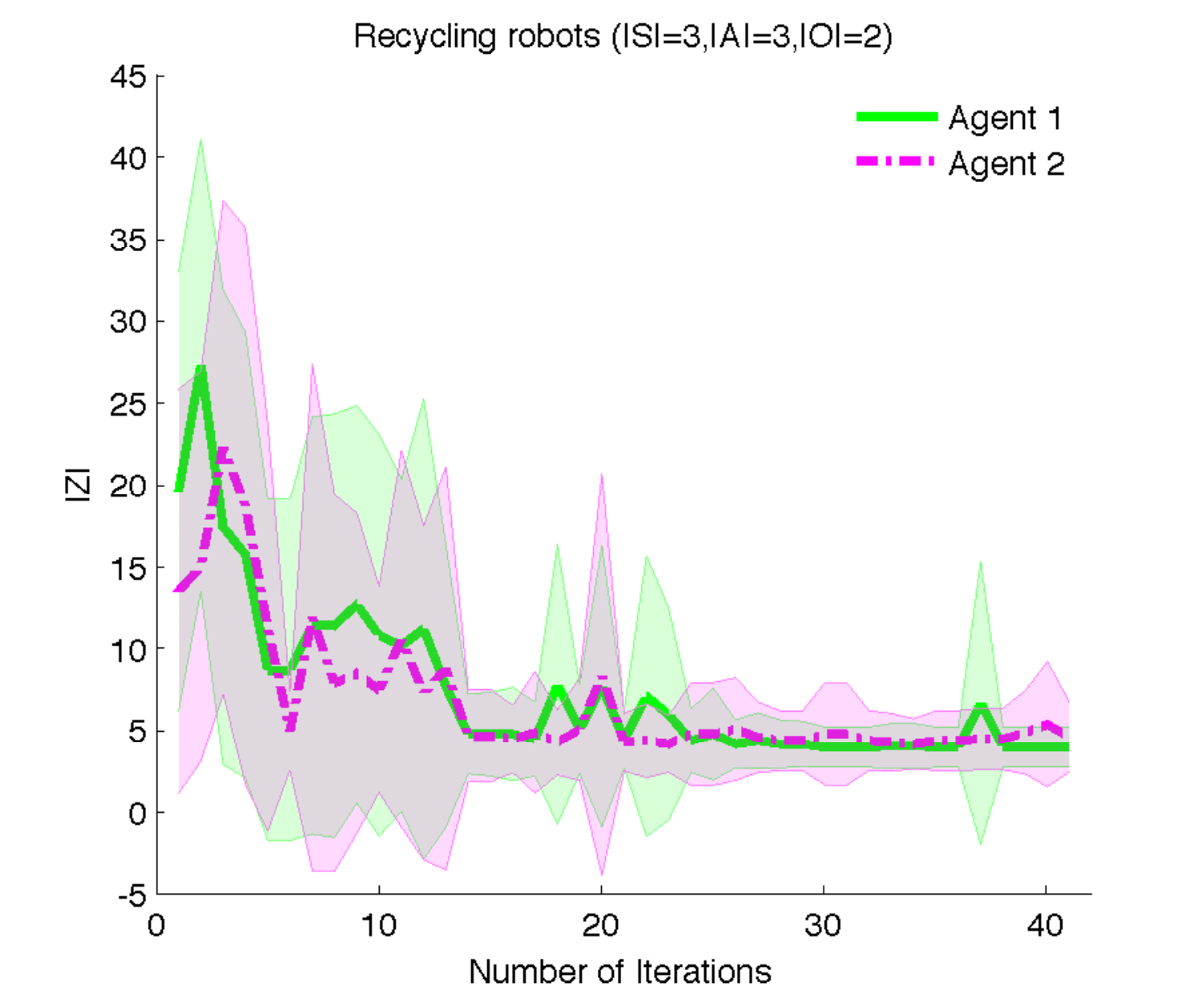}
\label{fig:SBtime}
}
\hspace{-0.75cm}
\subfigure{
\includegraphics[scale=0.35]{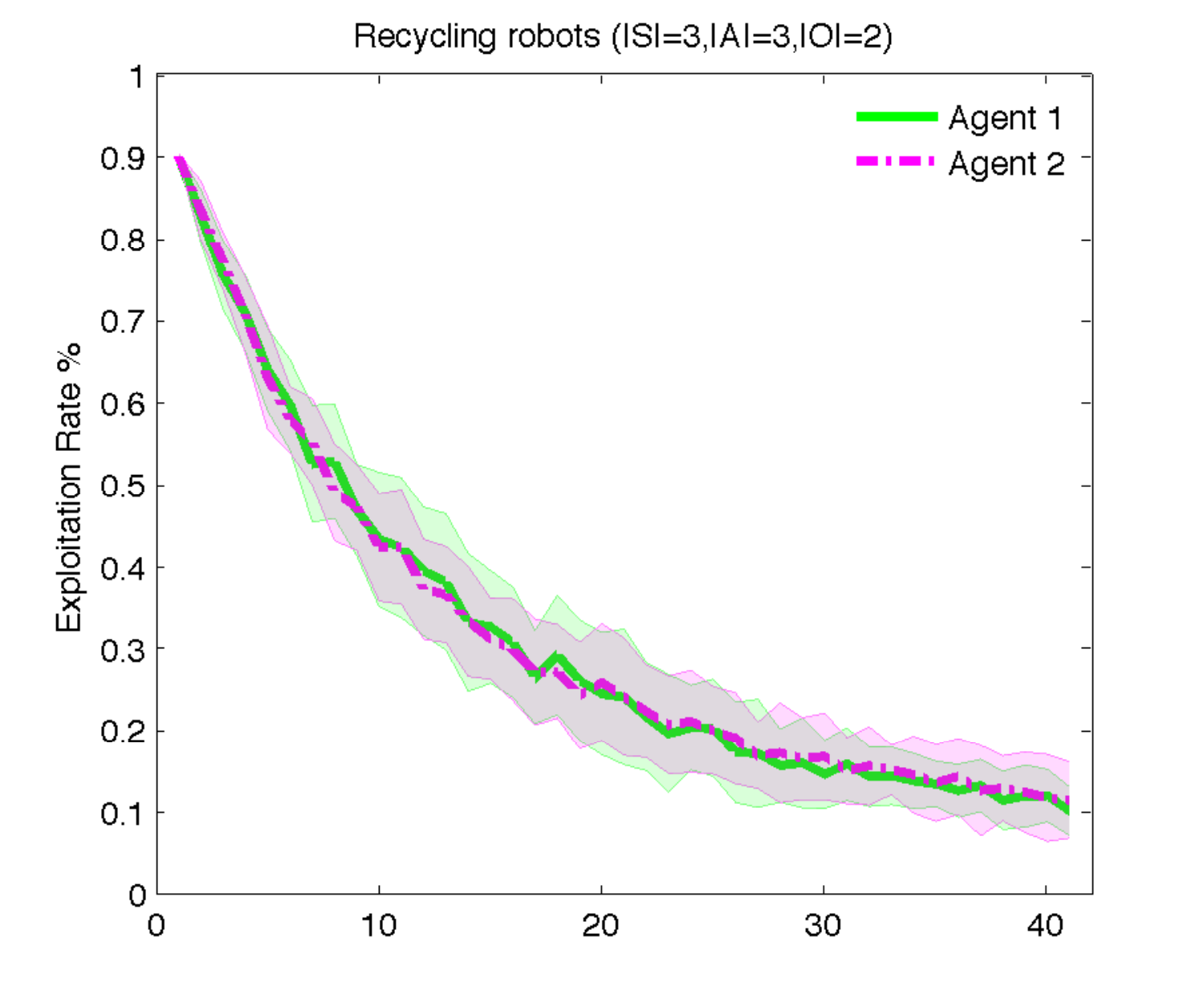}
\label{fig:SBtime}
}
\hspace{-0.75cm}
\subfigure{
\includegraphics[scale=0.35]{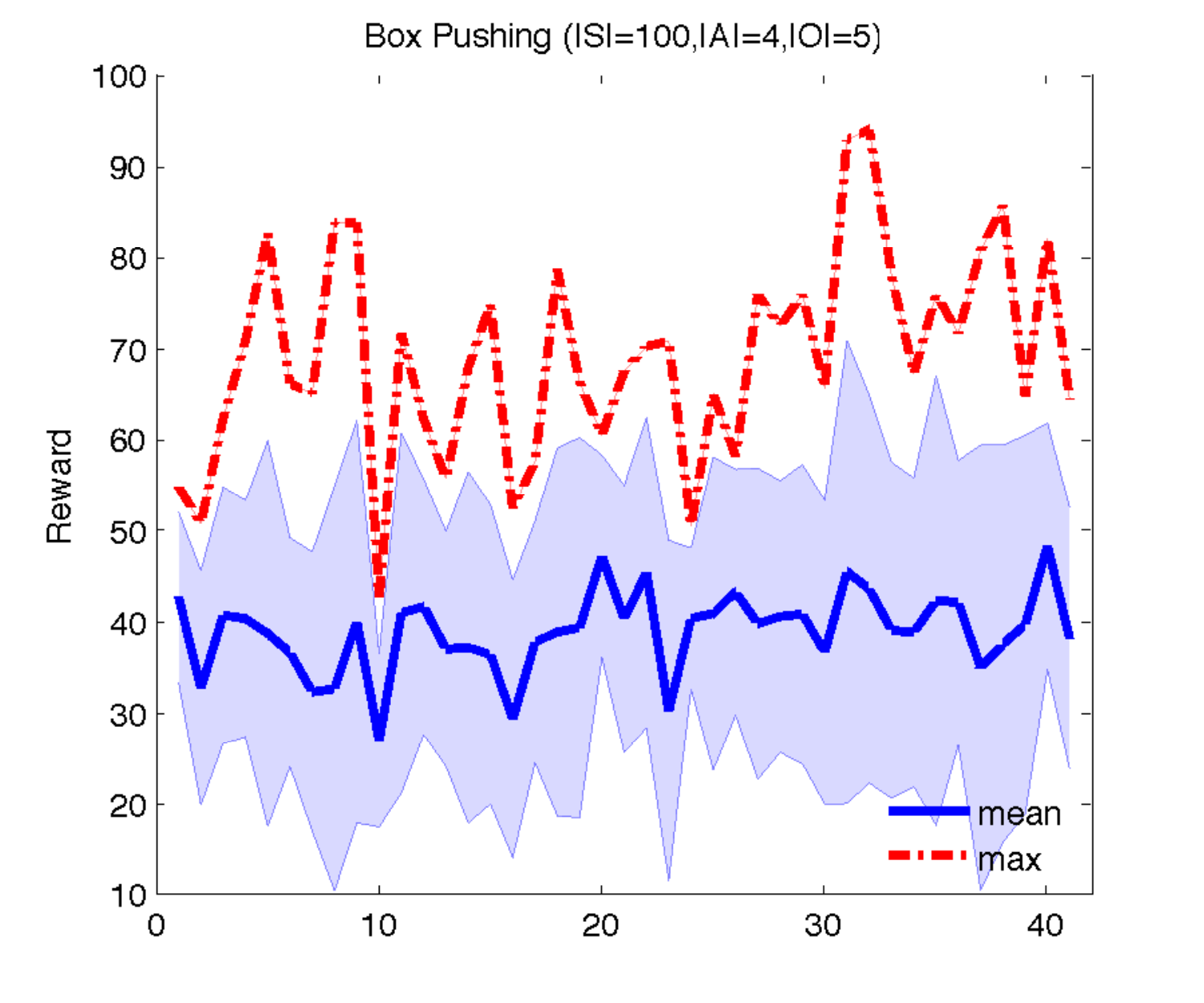}
\label{fig:EMtime}
}
\hspace{-0.75cm}
\subfigure{
\includegraphics[scale=0.35]{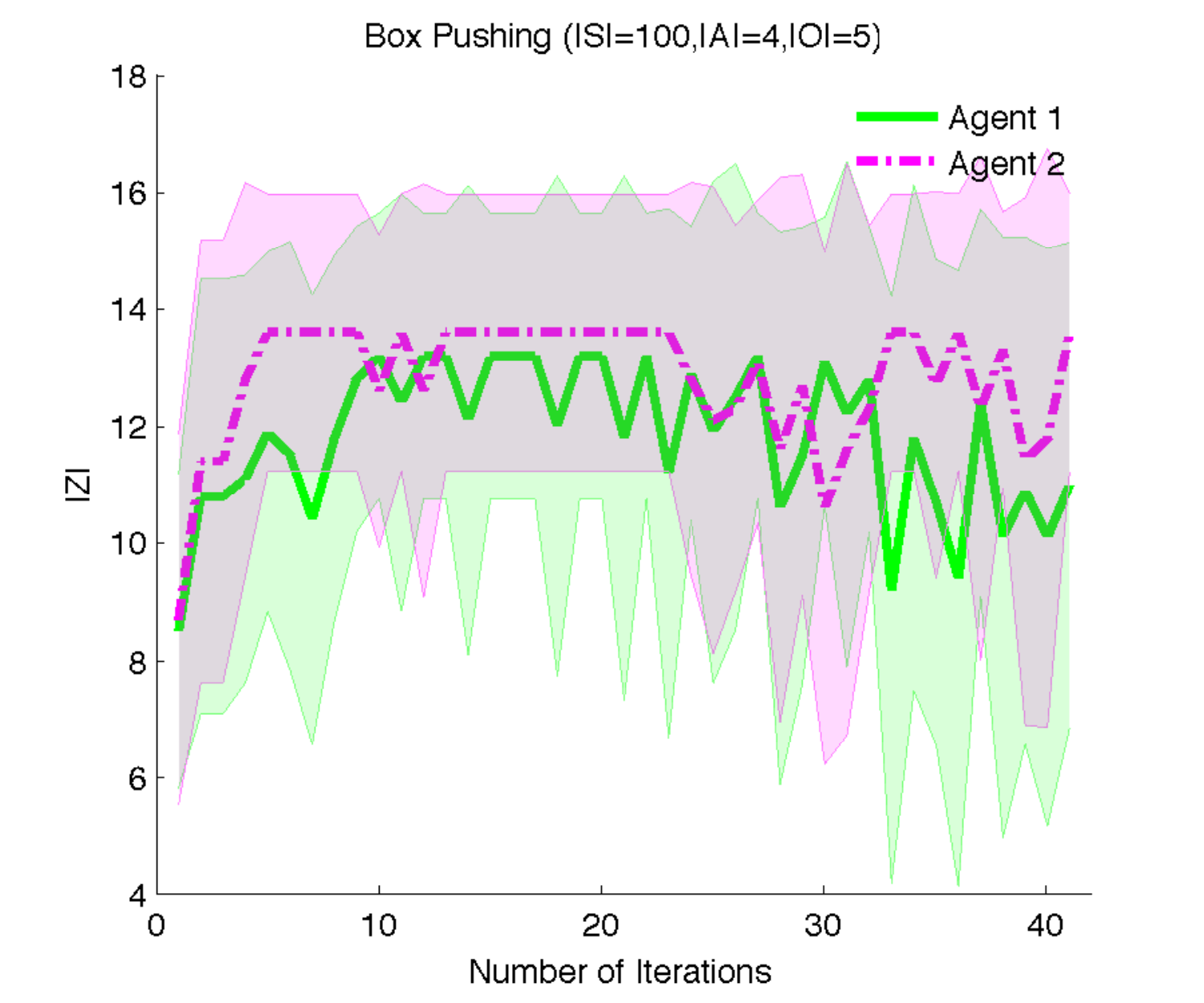}
\label{fig:SBtime}
}
\hspace{-0.75cm}
\subfigure{
\includegraphics[scale=0.35]{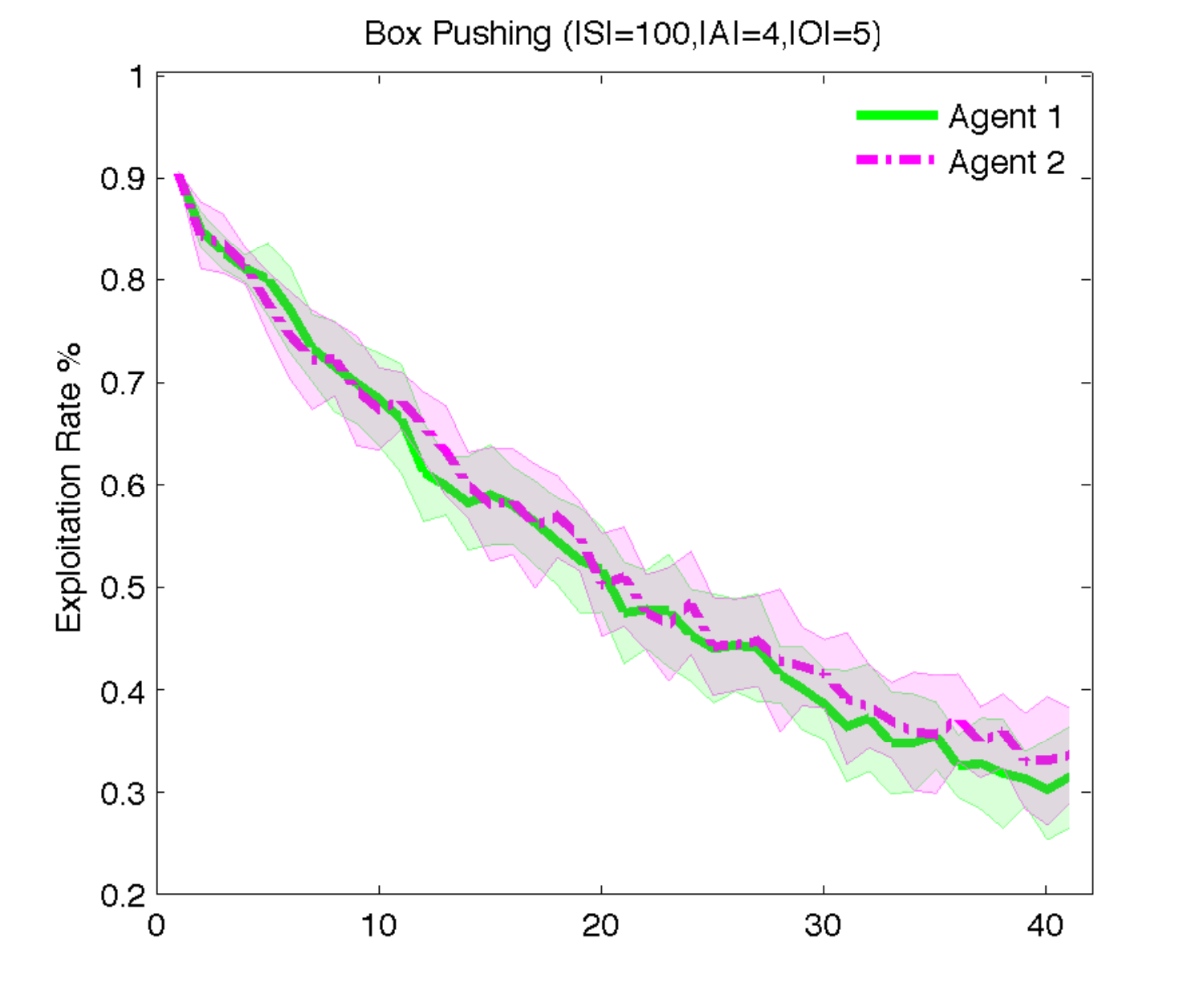}
\label{fig:SBtime}
}
\hspace{-0.75cm}
\subfigure{
\includegraphics[scale=0.35]{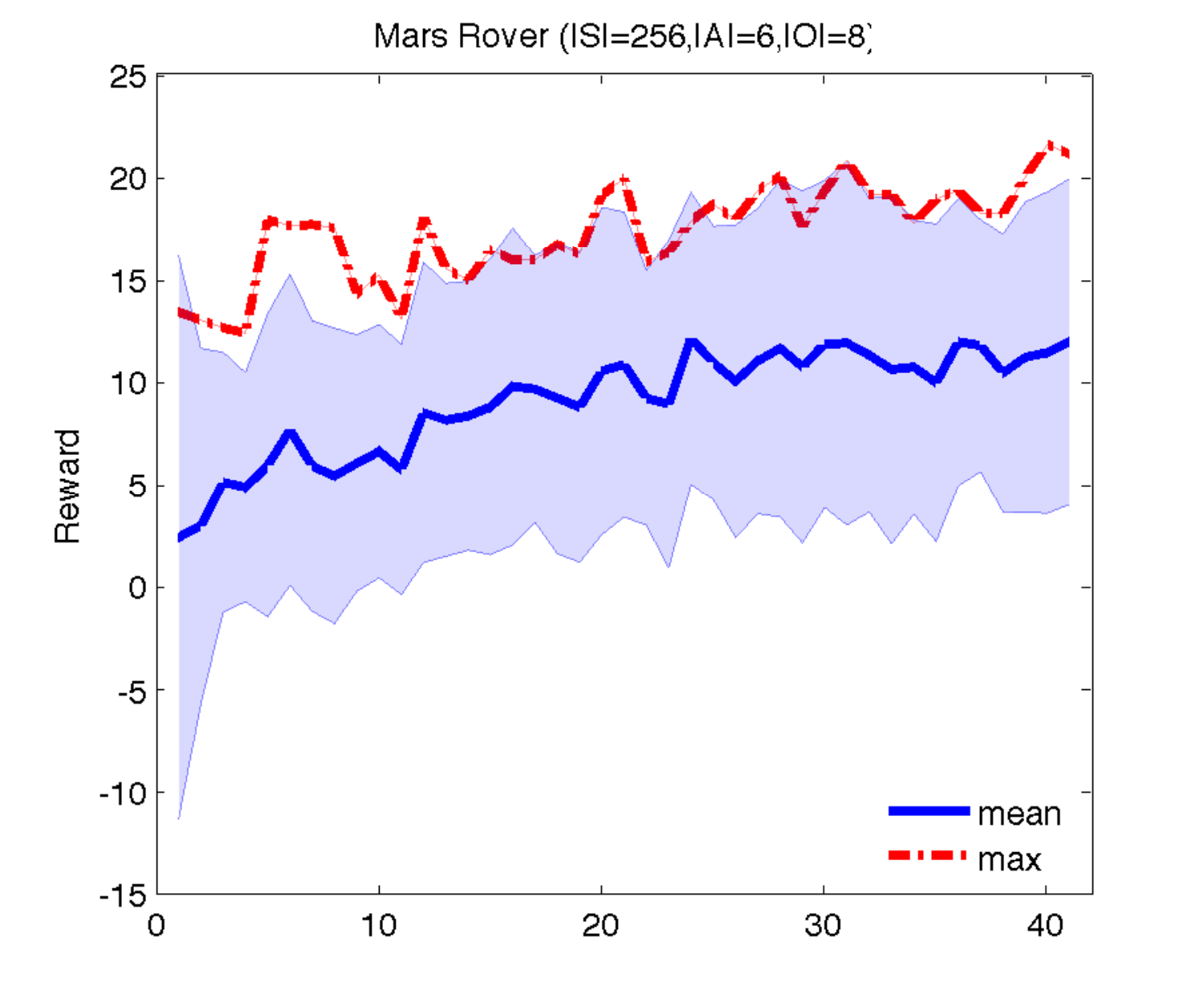}
\label{fig:EMtime}
}
\hspace{-0.75cm}
\subfigure{
\includegraphics[scale=0.35]{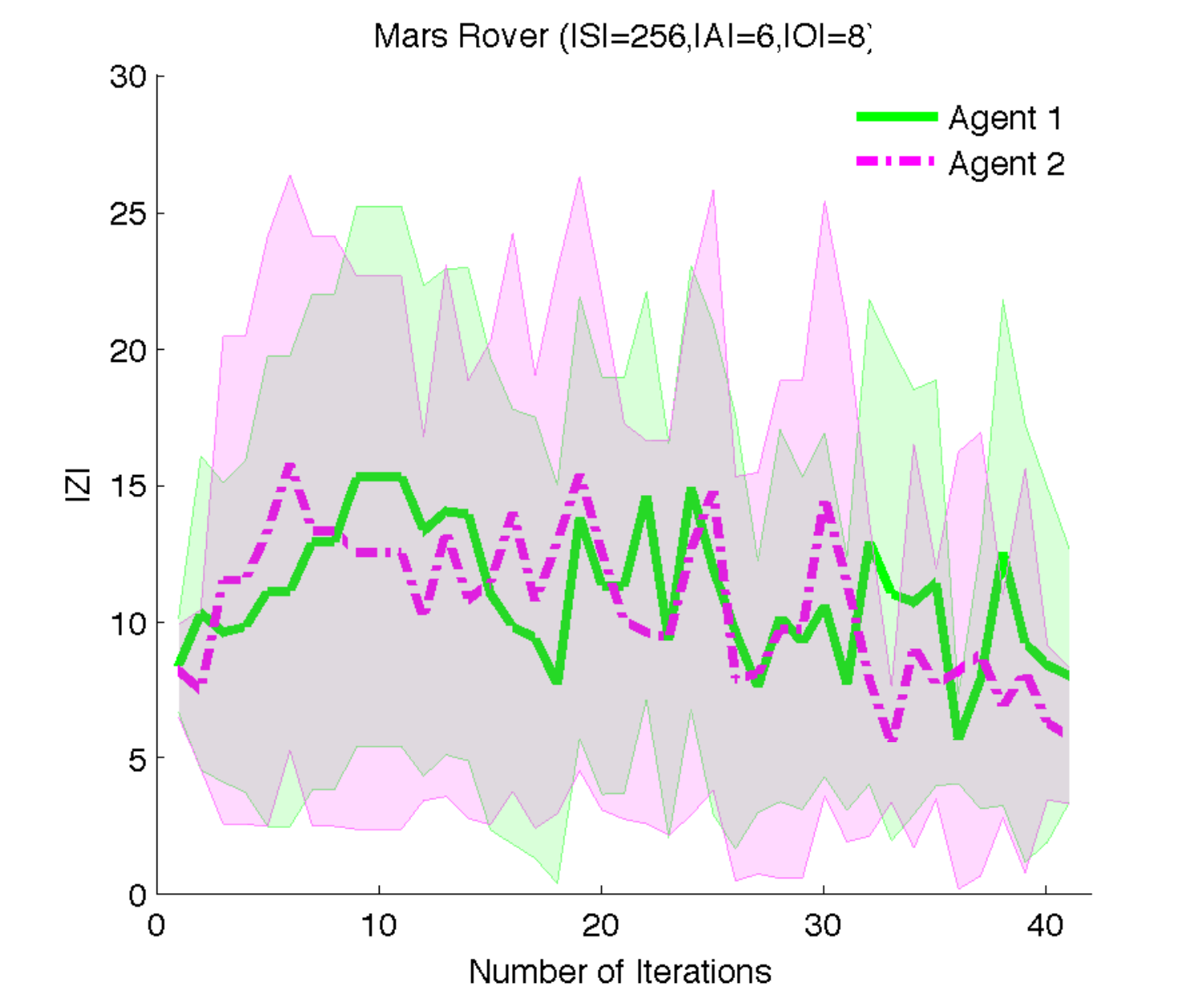}
\label{fig:SBtime}
}
\hspace{-0.75cm}
\subfigure{
\includegraphics[scale=0.35]{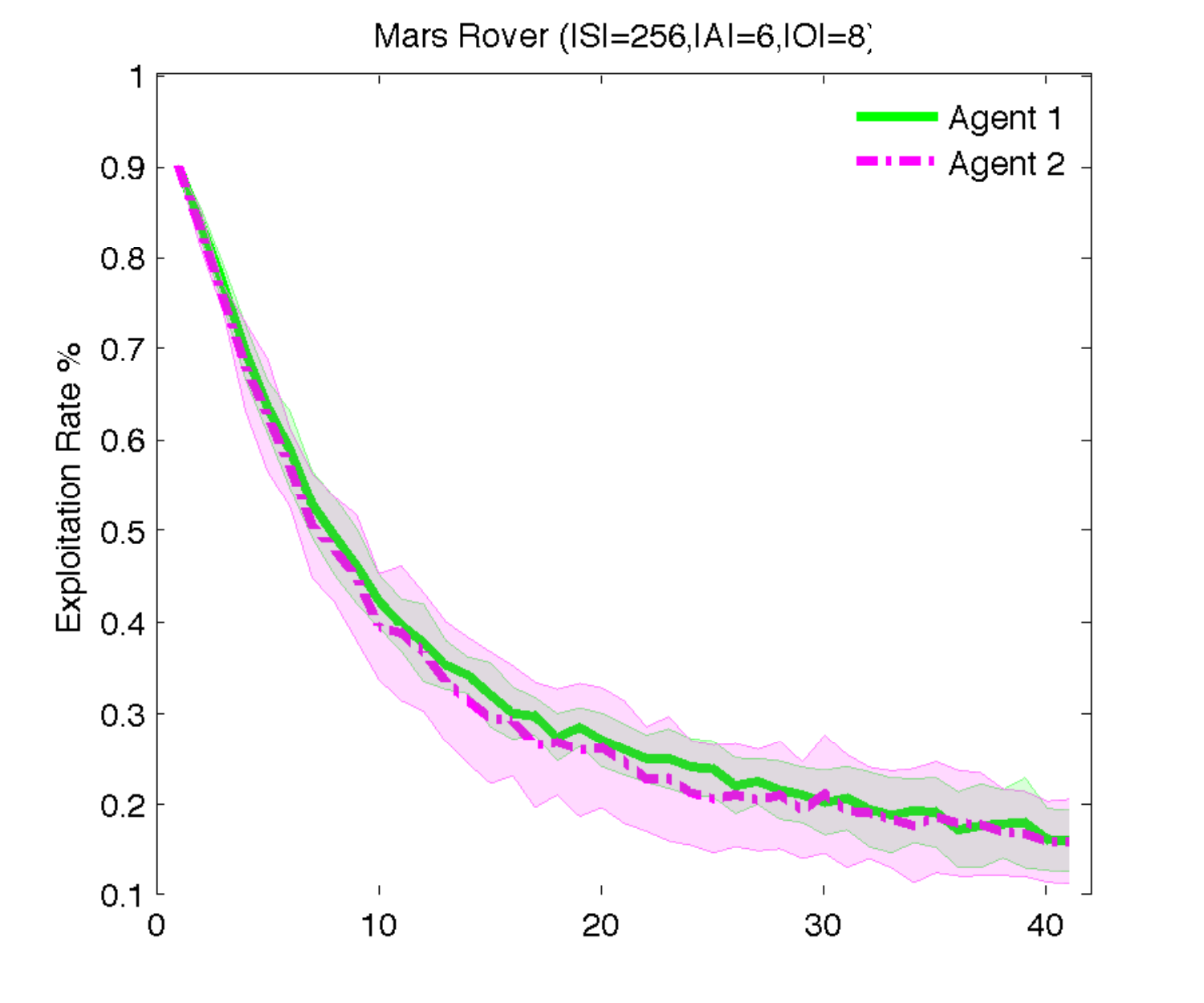}
\label{fig:SBtime}
}
\caption{Additional plots for illustrating exploration-exploitation tradeoff, including testing value (left), inferred controller numbers (middle) and exploration rate (right). In each iteration, a batch of samples are collected with updated behavior policies and are used to learn a set new policies with Algorithm~\ref{alg:Dec-SBPR-off-policy}.}
\label{fig:additional_plots2}
\end{figure}

Additional results from sequential batch learning with exploration exploitation trade-offs for five domains are plotted in Figure~\ref{fig:additional_plots2}. In each iteration, a batch of samples are collected with updated behavior policies\footnote{To generate these plots, 50 trajectories are collected in each iteration and the exploration parameter is set to be $u=100$.} and are used to learn a set new policies with Algorithm~\ref{alg:Dec-SBPR-off-policy}. These results are associated with the numbers reported in the first two columns of Table~\ref{tab:compare} in the main body of the paper. 
\begin{figure}[H]
\centering
\hspace{-0.2cm}
\subfigure{
\includegraphics[scale=0.35]{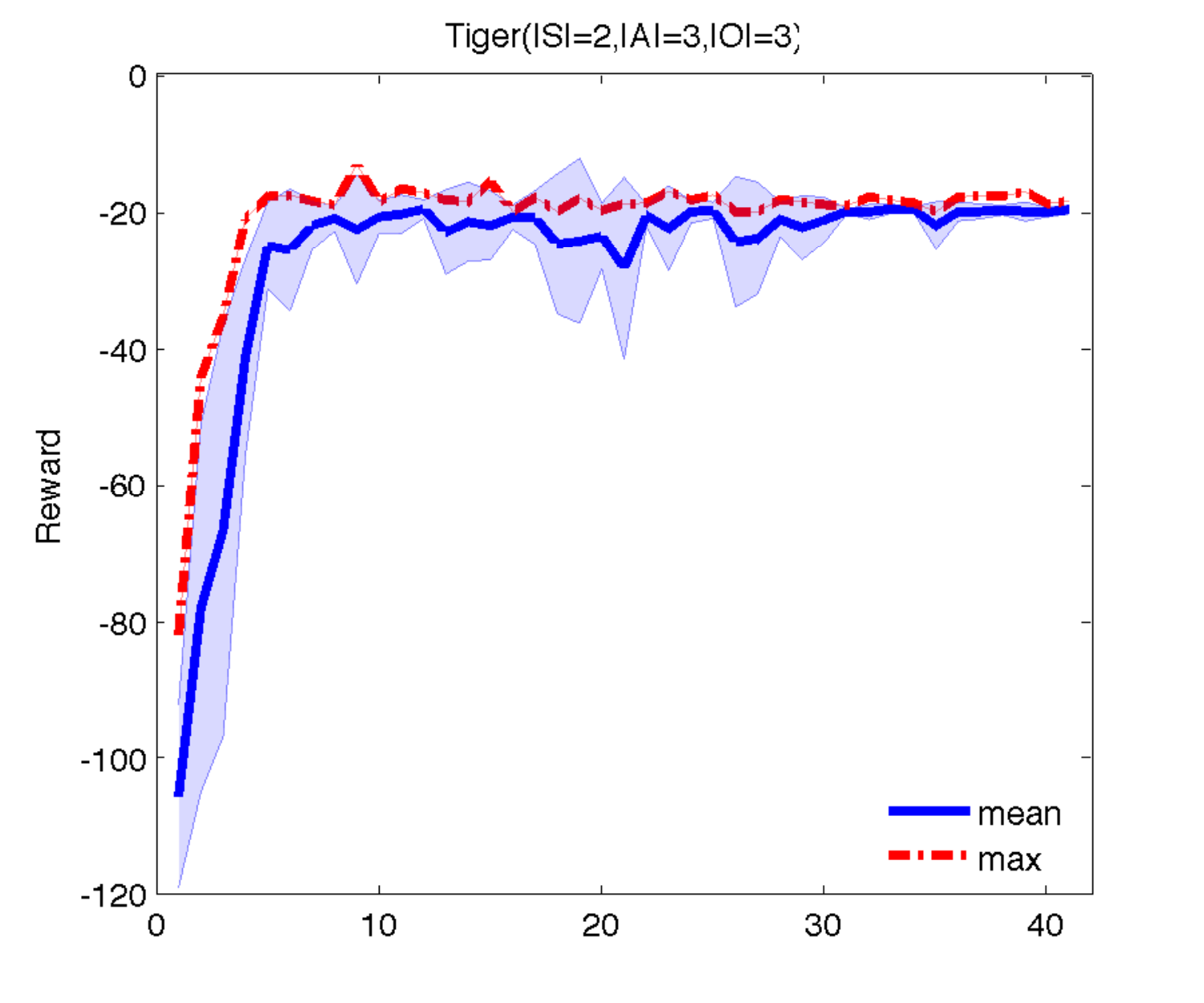}
\label{fig:RoverK30}
}
\hspace{-0.75cm}
\subfigure{
\includegraphics[scale=0.35]{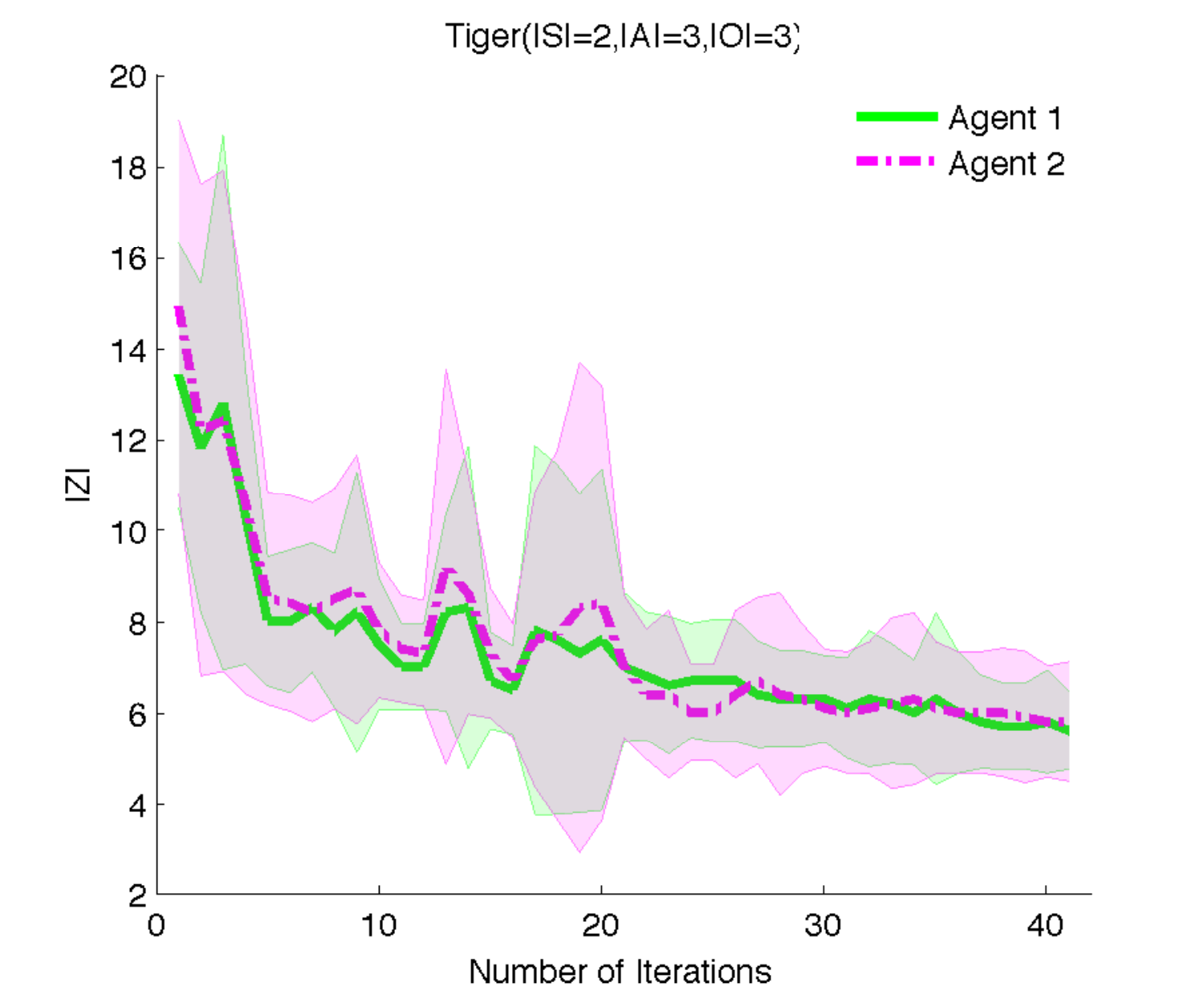}

\label{fig:RoverK100}
}
\hspace{-0.75cm}
\subfigure{
\includegraphics[scale=0.35]{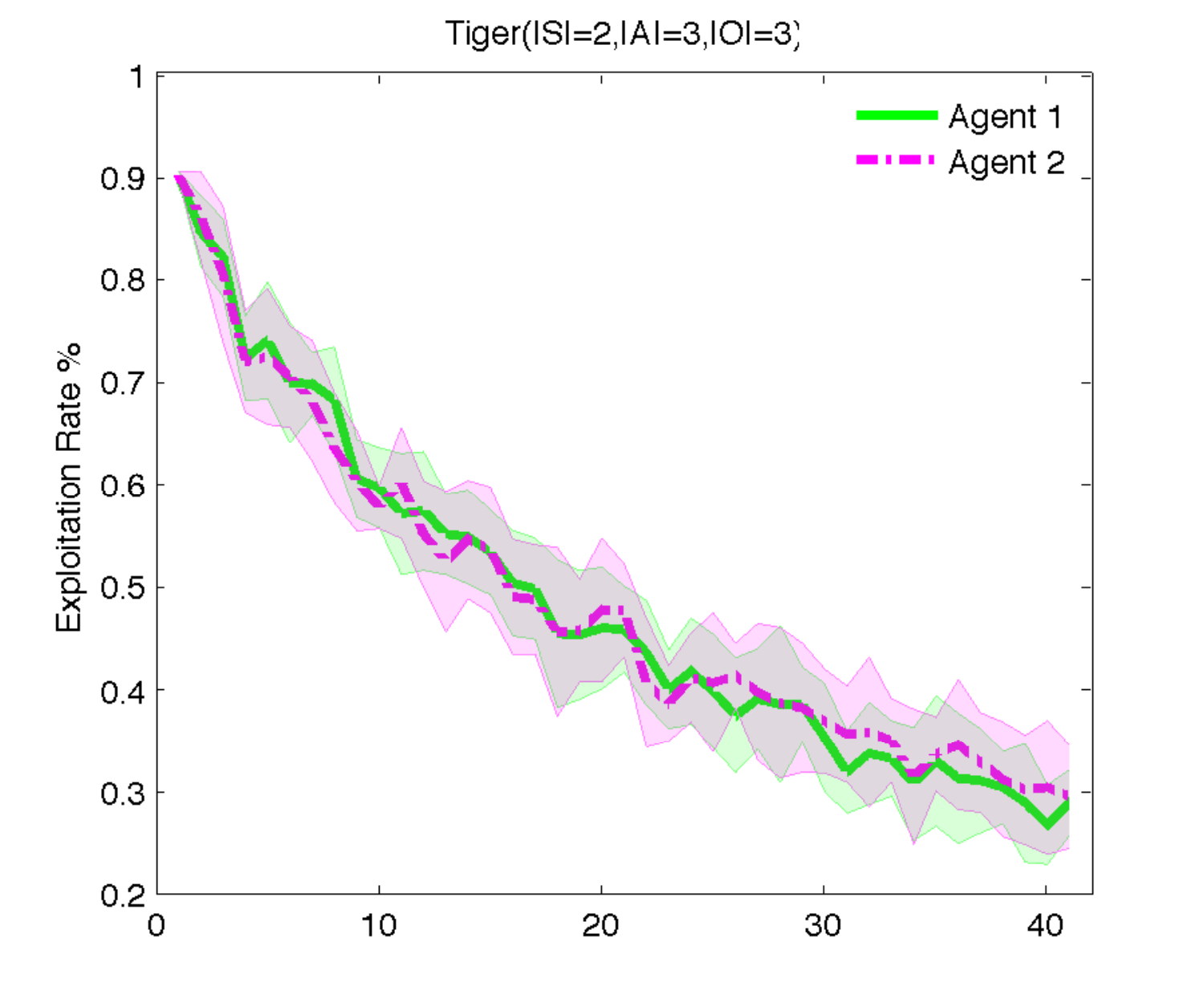}

\label{fig:RoverK300}
}
\end{figure}

\footnotesize
\bibliographystyle{natbib}
\bibliography{DecSBPR_arXiv}

\end{document}